\documentclass{article}

\PassOptionsToPackage{numbers,compress}{natbib}
\usepackage[main,final]{neurips_2026}

\usepackage[utf8]{inputenc}
\usepackage[T1]{fontenc}
\usepackage[table]{xcolor}
\usepackage{hyperref}
\usepackage{url}
\usepackage{booktabs}
\usepackage{amsfonts}
\usepackage{nicefrac}
\usepackage{microtype}
\usepackage{multirow}
\usepackage{graphicx}
\usepackage{amsmath}
\usepackage[normalem]{ulem}
\usepackage{soul}
\usepackage{placeins}
\hypersetup{
  hidelinks,
  pdftitle={MUSS: A Multi-scale and Sequence-based Model for Single-cell Gene Regulation},
  pdfauthor={Kangjie Zheng; Amirhossein Vahidi; Liying Jin; Joseph Clarke; Vijaya Baskar MS; Connor Rogerson; Mengji Zhang; Muzlifah Haniffa; Anshul Kundaje; Mohammad Lotfollahi},
  pdfsubject={NeurIPS 2026 camera-ready manuscript}
}

\definecolor{seenblue}{RGB}{221,235,247}
\definecolor{unseenpeach}{RGB}{248,222,212}
\definecolor{mygray}{gray}{.9}

\def\method{\textsc{Muss}}

\title{MUSS: A Multi-scale and Sequence-based Model for Single-cell Gene Regulation}

\author{%
  \textbf{Kangjie Zheng}$^{1}$\thanks{Equal contribution.}~~~
  \textbf{Amirhossein Vahidi}$^{1, 2}$${}^{*}$~~~
  \textbf{Liying Jin}$^{1}$${}^{*}$~~~
  \textbf{Joseph Clarke}$^{1}$ \\
  \textbf{Vijaya Baskar MS}$^{1}$~~~
  \textbf{Connor Rogerson}$^{1}$~~~
  \textbf{Mengji Zhang}$^{1}$ \\
  \textbf{Muzlifah Haniffa}$^{1,3,4}$~~~
  \textbf{Anshul Kundaje}$^{6,7}$~~~
  \textbf{Mo Lotfollahi}$^{1,2,5,8}$\thanks{Corresponding author: \texttt{ml19@sanger.ac.uk}.} \\
  \\
  $^{1}$Wellcome Sanger Institute, Wellcome Genome Campus, Cambridge, UK \\
  $^{2}$Cambridge Centre for AI in Medicine, University of Cambridge, Cambridge, UK \\
  $^{3}$Cambridge University Hospitals Foundation Trust, Addenbrooke's Hospital, Cambridge, UK \\
  $^{4}$Cambridge Institute for Therapeutic Immunology and Infectious Diseases (CIITID), \\
  University of Cambridge, UK \\
  $^{5}$Cambridge Stem Cell Institute, University of Cambridge, Cambridge, UK \\
  $^{6}$Department of Genetics, Stanford University School of Medicine, Stanford, CA, USA \\
  $^{7}$Department of Computer Science, Stanford University, Stanford, CA, USA \\
  $^{8}$Department of Medicine, University of Cambridge, Cambridge, UK \\
}

\begin{document}
\maketitle
\begin{abstract}
Modeling gene regulation in single cells requires jointly representing DNA sequence, chromatin accessibility, and gene expression across regulatory scales. Sequence-based models capture DNA-level regulatory grammar but often lack cell-level resolution, whereas many single-cell omics models represent genes and peaks as fixed-vocabulary tokens without their underlying sequence. Here, we present \method{}, a sequence-based multi-scale model for single-cell gene regulation that preserves both DNA-level and cell-level resolution. \method{} derives peak- and gene-level representations from DNA using a frozen pre-trained sequence encoder and integrates them through peak--peak self-attention and distance-biased peak--gene interactions. Peak masked language modeling followed by training on paired scATAC-seq and scRNA-seq aligns these representations with cell-specific regulatory states. Across gene expression prediction, enhancer--gene linking, motif-guided regulatory prioritization, and zero-shot cell clustering, \method{} achieves strong performance against sequence-based and single-cell omics baselines. It further predicts expression for genes and peaks unseen during task training. These results support sequence-grounded, multi-scale representations for decoding cell-type-specific regulatory programs. Code, datasets, and model checkpoints are available at \url{https://github.com/Lotfollahi-lab/MUSS_neurips26}.

\end{abstract}
\section{Introduction}
\label{sec::intro}
Cells share a genome yet express distinct transcriptional programs. Transcription factors (TFs) bind specific DNA motifs in accessible regulatory regions and influence target genes through combinatorial interactions spanning kilobases to megabases \citep{lambert2018human,schoenfelder2019long,gschwind2023encyclopedia}. Modeling this process requires both the sequence content of regulatory elements and the cell-specific chromatin state in which they operate. Paired single-cell chromatin accessibility (scATAC-seq) and expression (scRNA-seq) measurements provide an opportunity to connect these scales, while large-scale genomic and single-cell models supply transferable representations \citep{dalla-torre2024nucleotide,zhou2024dnabert2,nguyen2023hyenadna,cui2024scgpt, zheng2024esm}.

Existing approaches capture complementary aspects of regulation. Sequence-to-function models such as Borzoi and AlphaGenome predict functional readouts within genomic windows, while GET combines sequence-derived features with accessibility \citep{linder2025borzoi,avsec2026alphagenome,fu2025get}. These approaches do not directly integrate the genome-wide accessibility state of each individual cell with a shared representation of its peaks and genes. Recent work, notably scooby, does extend sequence-based prediction to single-cell resolution using a cell-specific decoder \citep{hingerl2025scooby}. In parallel, single-cell omics models learn cell-state-dependent gene or peak representations from expression and accessibility data \citep{theodoris2023geneformer,cui2024scgpt,hao2024scfoundation,yang2022scbert,jiao2025chromfound,chen2025epiagent}. However, fixed-vocabulary representations do not by themselves provide sequence-grounded embeddings for previously unseen regulatory entities. The remaining challenge is to unify DNA-derived peak and gene representations with cell-specific context across regulatory scales.

\begin{figure}[t]
    \centering
    \includegraphics[width=\textwidth]{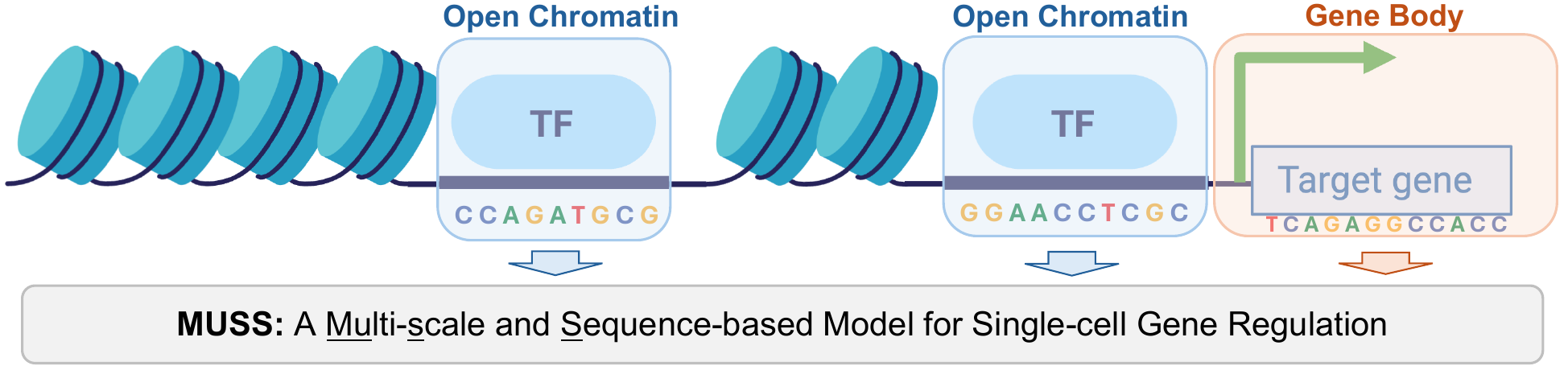}
    \caption{\method{} constructs unified sequence-based representations for chromatin-accessible regions (peaks) and genes to capture global, multi-scale regulatory semantics. The illustration incorporates artwork created with BioRender.}
    \label{fig::concept}
\end{figure}

We propose \method{} (Figure~\ref{fig::concept}), which derives both peak and gene embeddings from a frozen DNA sequence encoder. A multi-scale encoder models dependencies among accessible peaks through self-attention and exchanges information between peaks and genes through distance-biased cross-attention. Peak masked language modeling learns accessibility context, and subsequent supervision on paired scATAC-seq and scRNA-seq connects this context to gene expression. A genomic region absent from task training can still be encoded from its sequence and processed by the same interaction layers. Our contributions are:
\begin{itemize}
    \item \textbf{Unified multi-scale regulatory modeling.} A shared sequence-based representation links DNA, chromatin accessibility, and cell-state-dependent gene expression through peak--peak and peak--gene interactions.
    \item \textbf{Generalization beyond the training vocabulary.} Held-out-region experiments across four paired datasets demonstrate expression prediction for unseen genes and peaks, alongside transfer to unseen cell types.
    \item \textbf{Reusable regulatory representations.} Enhancer--gene linking, motif-guided prioritization, zero-shot clustering, and few-shot adaptation assess transfer beyond expression prediction. Additional baselines and cross-task ablations distinguish these gains from those of sequence embeddings alone.
\end{itemize}
Related work is discussed further in Appendix~\ref{sec::related_works}.

\section{Proposed Method: {\method}}
\label{sec:method}

\begin{figure*}[t]

    \centering
    \includegraphics[width=1.0\textwidth]{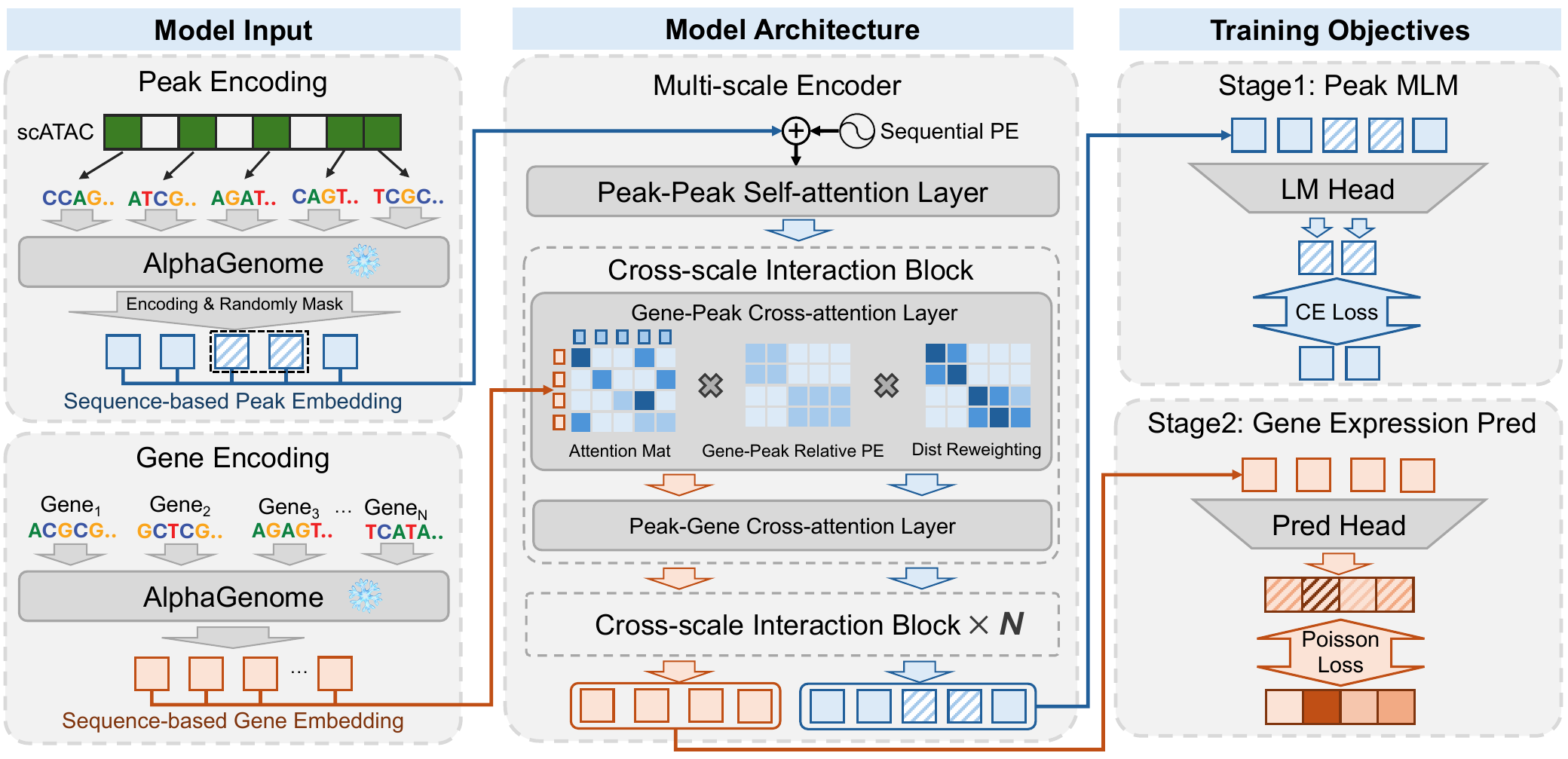}
    \caption{Overview of our proposed \method{}. \method{} first constructs sequence-based embeddings for peaks and genes using a frozen AlphaGenome encoder. It then models regulatory dependencies through a multi-scale encoder, including peak--peak self-attention and bidirectional cross-scale attention. Finally, \method{} is trained with peak masked language modeling and gene expression prediction to learn peak- and gene-level regulatory semantics.}
    \label{fig:main_framework}
\end{figure*}

\subsection{Model Overview}
\label{sec:model_overview}

For each cell, \method{} encodes accessible peaks and genes from DNA, integrates them with peak--peak self-attention and two-way peak--gene cross-attention, and learns representations using peak masked language modeling followed by gene expression prediction (Figure~\ref{fig:main_framework}). Genomic distance biases the cross-scale interactions.

\subsection{Unified Gene and Peak Sequence-based Encoding}
\label{sec:sequence_encoding}

To unify peak- and gene-level representations, \method{} encodes both entities directly from their genomic DNA sequences. Let $\mathcal{E}=\mathcal{P}\cup\mathcal{G}$ denote the union of peaks and genes, where each entity $e\in\mathcal{E}$ is associated with genomic coordinates $(c_e,s_e,t_e)$ on the human reference genome. We extract the corresponding sequence and encode it using a frozen AlphaGenome encoder $\Phi_{\mathrm{AG}}$:
\[
x_e=\mathrm{Seq}_{\mathrm{ref}}(c_e,s_e,t_e),
\qquad
H_e=\Phi_{\mathrm{AG}}(x_e)=[h_{e,1},\ldots,h_{e,L_e}].
\]
We use the 128-bp-resolution AlphaGenome embeddings and obtain the final entity representation by average pooling over all bins overlapping the interval:
\[
z_e=\frac{1}{L_e}\sum_{l=1}^{L_e}h_{e,l}.
\]
This produces sequence-based peak and gene embeddings,
\[
Z^p=\{z_{p_i}\}_{i=1}^{N_p},
\qquad
Z^g=\{z_{g_j}\}_{j=1}^{N_g},
\]
which are initialized in a shared DNA-grounded semantic space because the same frozen encoder is used for both scales.

For the first-stage peak masked language modeling (MLM) task, we randomly mask 15\% of peaks by replacing their embeddings with a learnable \texttt{[MASK]} embedding:
\[
\tilde z_{p_i}=
\begin{cases}
z_{\mathrm{mask}}, & p_i\in\mathcal{M},\\
z_{p_i}, & p_i\notin\mathcal{M},
\end{cases}
\qquad |\mathcal{M}|=0.15N_p .
\]
This gives the corrupted peak embedding set \(\tilde Z^p=\{\tilde z_{p_i}\}_{i=1}^{N_p}\), which is then used as input for peak-level pretraining.

\subsection{Multi-scale Encoder}
\label{sec:multi_scale_encoder}

The multi-scale encoder consists of a peak--peak self-attention layer and stacked cross-scale interaction blocks.

\paragraph{Peak--Peak Self-attention.}
Given peak embeddings $Z^p=\{z_{p_i}\}_{i=1}^{N_p}$, we apply full self-attention \cite{vaswani2017attention} to model global dependencies among accessible regions. We compute
\[
Q^p=Z^pW_Q^p,\quad K^p=Z^pW_K^p,\quad V^p=Z^pW_V^p,
\]
and then apply RoPE~\cite{su2024roformer} to $Q^p$ and $K^p$ using the peak order as positional indices. The updated representations are
\[
\hat Z^p
=
\mathrm{softmax}\left(
\frac{\mathrm{RoPE}(Q^p)\mathrm{RoPE}(K^p)^\top}{\sqrt{d}}
\right)V^p,
\]
which are then fed into cross-scale interaction blocks.

\paragraph{Cross-scale Interaction.}
Each block exchanges information between peaks and genes via bidirectional cross-attention. Gene-to-peak attention aggregates regulatory signals to update gene embeddings, while peak-to-gene attention refines peak representations using gene context:
\[
Z^g \leftarrow Z^g + \mathrm{CrossAttn}(Z^g,\hat Z^p;B^{gp}),\quad
\hat Z^p \leftarrow \hat Z^p + \mathrm{CrossAttn}(\hat Z^p,Z^g;B^{pg}),
\]
followed by feed-forward layers with residual connections. Here $B^{pg}\in\mathbb{R}^{H\times N_p\times N_g}$ is the peak-to-gene bias, and $B^{gp}$ is its transpose. Cross-attention is defined as
\[
\mathrm{CrossAttn}(X_q,X_{kv};B)
=
\mathrm{softmax}\left(
\frac{(X_qW_Q)(X_{kv}W_K)^\top}{\sqrt{d_h}}+B
\right)X_{kv}W_V.
\]

\paragraph{Genomic Relative Positional Bias.}
We encode peak--gene genomic distance as an additive attention bias. For intra-chromosomal pairs with signed distance $d_{ij}$, we transform the distance magnitude using Gaussian RBFs \cite{scholkopf1997comparing} over $\log(1+|d_{ij}|)$, followed by a nonlinear projection:
\[
B^{\mathrm{rpe}}_{:,i,j}
= f_{\theta}\big(\mathrm{RBF}(\log(1+|d_{ij}|))\big)=
f_{\theta}\left(
\left[
\exp\left\{-\gamma_k
\left(\log(1+|d_{ij}|)-\mu_k\right)^2
\right\}
\right]_{k=1}^{K}
\right),
\]
where $\mu_k$ and $\gamma_k>0$ are learnable kernel centers and bandwidths. The projection $f_{\theta}$ maps RBF features to per-head biases. For inter-chromosomal pairs, we assign a constant bias $B^{\mathrm{rpe}}_{:,i,j}=\beta_{\mathrm{inter}}$.
\paragraph{Distance-based Reweighting.}
To reflect the decay of regulatory effects with distance~\cite{fulco2019activity} and avoid overly diffuse attention, we add a monotonic penalty for intra-chromosomal pairs:
\[
B^{\mathrm{dist}}_{:,i,j}
=
-\lambda \log(1+|d_{ij}|),
\]
where $\lambda=\mathrm{softplus}(\lambda_0)\ge 0$ is a learnable non-negative decay parameter. The final bias is
\[
B^{pg}=B^{\mathrm{rpe}}+B^{\mathrm{dist}},
\]
with $B^{gp}$ obtained by transposing the peak and gene dimensions of $B^{pg}$.
\subsection{Model Training}
\label{sec:model_training}

To provide supervision at both regulatory scales, \method{} is trained with two complementary objectives. The first stage supervises peak-level representations through masked peak prediction, while the second stage supervises gene-level representations through gene expression prediction.

\paragraph{Stage 1: Peak MLM Training.}For peak-level pretraining, we use the masked peak inputs $\tilde Z^p$ defined in Section~\ref{sec:sequence_encoding}. After passing them through the multi-scale encoder, we obtain contextualized peak representations $\bar Z^p=\{\bar z_{p_i}\}_{i=1}^{N_p}$. For each masked peak $p_i\in\mathcal{M}$, an LM head predicts its original peak identity:
\[
\ell_i = \bar z_{p_i} E_p^\top,
\qquad
\hat y_i=\mathrm{softmax}(\ell_i),
\]

where $E_p \in \mathbb{R}^{|\mathcal{V}_p| \times d}$ denotes the pre-trained sequence embedding matrix for the training peak vocabulary $\mathcal{V}_p$, not only the $N_p$ peaks in one cell. It is tied to the LM head. The peak MLM objective is:

\[
\mathcal{L}_{\mathrm{peak}}
=
-\frac{1}{|\mathcal{M}|}
\sum_{p_i\in\mathcal{M}}
\log \hat y_i[p_i].
\]

\paragraph{Stage 2: Gene Expression Training.}In the second stage, we supervise gene-level representations with gene expression prediction. Given the contextualized gene representations $\bar Z^g=\{\bar z_{g_j}\}_{j=1}^{N_g}$ from the multi-scale encoder, a prediction head estimates the expression rate of each gene:
\[
\hat \mu_j = f_{\mathrm{expr}}(\bar z_{g_j}),
\]
where $\hat \mu_j$ denotes the predicted expression mean for gene $g_j$. We train the model using a Poisson loss against the observed gene expression count $y_j$:
\[
\mathcal{L}_{\mathrm{gene}}
=
\frac{1}{N_g}
\sum_{j=1}^{N_g}
\left(
\hat \mu_j - y_j \log \hat \mu_j
\right).
\]
This objective guides \method{} to learn gene-level representations that are predictive of cell-state-dependent transcriptional outputs.

\paragraph{Generalization to unseen peaks.}
The training softmax uses $\mathcal{V}_p$, but unseen-peak representations come from the frozen sequence encoder, not from the classifier. New peaks are encoded from DNA and processed by the same distance-aware interaction layers. The tied MLM head can score an additional peak by appending its sequence embedding to $E_p$; this does not constitute supervision on that peak. The MLM head is discarded for downstream inference. Thus, downstream prediction is not restricted to the training peak vocabulary.

\section{Experimental Results}
\label{sec:exp}
\begin{table*}[t]
\centering
\setlength{\tabcolsep}{3.5pt}
\renewcommand{\arraystretch}{1.12}
\caption{\textit{Gene expression prediction from scATAC-seq to scRNA-seq under four evaluation settings defined by sequence split (\textbf{seen seq} vs.\ \textbf{unseen seq}) and cell-type split (\textbf{seen cell types} vs.\ \textbf{unseen cell types}). Higher is better for PCC, CCC, and $R^2$. Metrics are computed on per-cell-type pseudobulk. Best and second-best values are bold and underlined, respectively.}}
\label{tab:cross_omics_prediction_main}

\resizebox{\textwidth}{!}{
\begin{tabular}{ll|ccc|ccc|ccc|ccc}
\toprule
\multicolumn{2}{c|}{}
& \multicolumn{3}{c|}{\textbf{PBMC10K}~\citep{zheng2017massively}}
& \multicolumn{3}{c|}{\textbf{BMMC~\cite{luecken2021sandbox}}}
& \multicolumn{3}{c|}{\textbf{Openproblem~\cite{open-problems-multimodal}}}
& \multicolumn{3}{c}{\textbf{Bone~\cite{to2024multi}}} \\
\cmidrule(lr){3-5}\cmidrule(lr){6-8}\cmidrule(lr){9-11}\cmidrule(lr){12-14}
\multicolumn{1}{c}{} & \textbf{Model}
& PCC$\uparrow$ & CCC$\uparrow$ & $R^2\uparrow$
& PCC$\uparrow$ & CCC$\uparrow$ & $R^2\uparrow$
& PCC$\uparrow$ & CCC$\uparrow$ & $R^2\uparrow$
& PCC$\uparrow$ & CCC$\uparrow$ & $R^2\uparrow$ \\
\midrule

\rowcolor{unseenpeach}
\multicolumn{14}{c}{\textbf{Unseen sequence}} \\
\multicolumn{14}{l}{\textbf{\ul{Seen cell types.}}} \\

\multirow{3}{*}{\textit{Seq}}
& Borzoi~\cite{linder2025borzoi}                               & 0.560 & 0.237 & -0.293 & 0.539 & 0.424 & 0.025 & 0.613 & 0.486 & 0.242 & 0.487 & 0.313 & -0.040 \\
& AlphaGenome~\cite{avsec2026alphagenome}                          & \underline{0.609} & \underline{0.402} & \underline{-0.005} & 0.650 & 0.450 & 0.144 & \textbf{0.774} & 0.690 & \underline{0.522} & 0.615 & 0.545 & 0.262 \\
& scooby~\cite{hingerl2025scooby} & 0.533 & 0.353 & -0.088 & \underline{0.655} & \underline{0.543} & \underline{0.293} & 0.709 & \underline{0.696} & 0.312 & \underline{0.655} & \textbf{0.648} & \underline{0.278} \\
\cmidrule(lr){1-14}
\multirow{2}{*}{\textit{ATAC+Seq}}
& GET~\cite{fu2025get}                                  & 0.310 & 0.285 & -0.094 & 0.216 & 0.180 & -0.774 & 0.206 & 0.188 & -0.208 & 0.203 & 0.103 & -2.440 \\
& EpiBERT~\cite{javed2025epibert}                              & 0.491 & 0.386 & -0.063 & 0.549 & 0.540 & 0.098 & 0.595 & 0.571 & 0.310 & 0.475 & 0.448 & 0.040 \\
\cmidrule(lr){1-14}
\rowcolor{mygray}
& \method{}                            & \textbf{0.684} & \textbf{0.663} & \textbf{0.409} & \textbf{0.748} & \textbf{0.738} & \textbf{0.524} & \underline{0.755} & \textbf{0.735} & \textbf{0.552} & \textbf{0.683} & \underline{0.611} & \textbf{0.427} \\

\midrule
\multicolumn{14}{l}{\textbf{\ul{Unseen cell types.}}} \\

\multirow{3}{*}{\textit{Seq}}
& Borzoi~\cite{linder2025borzoi}                               & 0.571 & 0.303 & -0.134 & 0.549 & 0.457 & -0.011 & 0.606 & 0.469 & 0.227 & 0.483 & 0.315 & -0.040 \\
& AlphaGenome~\cite{avsec2026alphagenome}                          & 0.613 & 0.425 & 0.057 & 0.617 & 0.370 & -0.052 & \textbf{0.768} & \underline{0.676} & \underline{0.506} & 0.591 & 0.519 & 0.221 \\
& scooby~\cite{hingerl2025scooby} & 0.439 & 0.319 & -0.152 & \underline{0.628} & 0.536 & \underline{0.232} & 0.681 & 0.659 & 0.305 & \underline{0.655} & \textbf{0.648} & \underline{0.231} \\
\cmidrule(lr){1-14}
\multirow{2}{*}{\textit{ATAC+Seq}}
& GET~\cite{fu2025get}                                  & 0.342 & 0.298 & -0.187 & 0.196 & 0.168 & -0.752 & 0.215 & 0.196 & -0.208 & 0.201 & 0.110 & -2.402 \\
& EpiBERT~\cite{javed2025epibert}                              & \underline{0.639} & \underline{0.587} & \underline{0.304} & 0.564 & \underline{0.550} & 0.105 & 0.618 & 0.606 & 0.340 & 0.548 & 0.517 & 0.150 \\
\cmidrule(lr){1-14}
\rowcolor{mygray}
& \method{}                            & \textbf{0.698} & \textbf{0.686} & \textbf{0.436} & \textbf{0.742} & \textbf{0.724} & \textbf{0.481} & \underline{0.762} & \textbf{0.742} & \textbf{0.564} & \textbf{0.683} & \underline{0.609} & \textbf{0.419} \\

\midrule
\rowcolor{seenblue}
\multicolumn{14}{c}{\textbf{Seen sequence}} \\
\multicolumn{14}{l}{\textbf{\ul{Seen cell types.}}} \\

\multirow{4}{*}{\textit{Non-seq}}
& CMAE~\cite{yang2021cmae}             & 0.535 & 0.394 & -0.255 & 0.441 & 0.298 & -0.639 & 0.679 & 0.483 & -0.844 & 0.882 & 0.626 & -0.586 \\
& scMoGNN~\cite{wen2022scmognn}        & \underline{0.851} & 0.808 & \underline{0.665} & \underline{0.891} & 0.847 & 0.730 & \textbf{0.996} & \textbf{0.995} & \textbf{0.991} & \underline{0.955} & \textbf{0.945} & \underline{0.896} \\
& ChromFound~\cite{jiao2025chromfound} & 0.751 & 0.514 & 0.078 & 0.855 & 0.620 & 0.264 & 0.879 & 0.690 & 0.148 & 0.919 & 0.646 & -0.248 \\
& EpiAgent~\cite{chen2025epiagent} & 0.763 & 0.524 & 0.092 & 0.819 & 0.575 & 0.131 & 0.876 & 0.683 & 0.116 & 0.929 & 0.666 & -0.137 \\
\cmidrule(lr){1-14}
\multirow{3}{*}{\textit{Seq}}
& Borzoi~\cite{linder2025borzoi}                               & 0.716 & 0.458 & 0.075 & 0.716 & 0.617 & 0.345 & 0.795 & 0.707 & 0.537 & 0.672 & 0.462 & 0.151 \\
& AlphaGenome~\cite{avsec2026alphagenome}                          & 0.653 & 0.434 & -0.049 & 0.691 & 0.498 & 0.160 & 0.841 & 0.794 & 0.660 & 0.643 & 0.458 & 0.081 \\
& scooby~\cite{hingerl2025scooby} & 0.829 & 0.809 & 0.648 & 0.747 & 0.620 & 0.397 & \underline{0.928} & \underline{0.928} & 0.853 & 0.815 & 0.793 & 0.610 \\
\cmidrule(lr){1-14}
\multirow{2}{*}{\textit{ATAC+Seq}}
& GET~\cite{fu2025get}                                  & 0.832 & \underline{0.810} & 0.638 & 0.608 & 0.602 & 0.213 & 0.925 & 0.903 & 0.799 & 0.868 & \underline{0.854} & 0.700 \\
& EpiBERT~\cite{javed2025epibert}                              & 0.772 & 0.736 & 0.588 & 0.880 & \underline{0.875} & \underline{0.755} & 0.815 & 0.807 & 0.661 & 0.835 & 0.820 & 0.616 \\
\cmidrule(lr){1-14}
\rowcolor{mygray}
& \method{}                            & \textbf{0.909} & \textbf{0.895} & \textbf{0.811} & \textbf{0.983} & \textbf{0.981} & \textbf{0.962} & \textbf{0.996} & \textbf{0.995} & \underline{0.990} & \textbf{0.958} & \textbf{0.945} & \textbf{0.897} \\

\midrule
\multicolumn{14}{l}{\textbf{\ul{Unseen cell types.}}} \\

\multirow{4}{*}{\textit{Non-seq}}
& CMAE~\cite{yang2021cmae}             & 0.520 & 0.370 & -0.430 & 0.432 & 0.304 & -0.586 & 0.669 & 0.480 & -0.862 & 0.881 & 0.645 & -0.464 \\
& scMoGNN~\cite{wen2022scmognn}        & 0.819 & 0.794 & 0.631 & \underline{0.856} & 0.794 & 0.613 & \underline{0.975} & \underline{0.974} & \underline{0.949} & \underline{0.937} & \textbf{0.928} & \textbf{0.863} \\
& ChromFound~\cite{jiao2025chromfound} & 0.723 & 0.473 & -0.102 & 0.815 & 0.614 & 0.264 & 0.855 & 0.662 & 0.062 & 0.908 & 0.654 & -0.204 \\
& EpiAgent~\cite{chen2025epiagent} & 0.746 & 0.503 & -0.009 & 0.818 & 0.620 & 0.248 & 0.860 & 0.669 & 0.079 & 0.915 & 0.680 & -0.070 \\
\cmidrule(lr){1-14}
\multirow{3}{*}{\textit{Seq}}
& Borzoi~\cite{linder2025borzoi}                               & 0.711 & 0.515 & 0.186 & 0.702 & 0.623 & 0.283 & 0.787 & 0.696 & 0.525 & 0.662 & 0.463 & 0.139 \\
& AlphaGenome~\cite{avsec2026alphagenome}                          & 0.650 & 0.436 & -0.035 & 0.645 & 0.444 & 0.007 & 0.828 & 0.782 & 0.642 & 0.626 & 0.420 & 0.022 \\
& scooby~\cite{hingerl2025scooby} & 0.773 & 0.766 & 0.550 & 0.711 & 0.646 & 0.401 & 0.887 & 0.884 & 0.783 & 0.788 & 0.764 & 0.540 \\
\cmidrule(lr){1-14}
\multirow{2}{*}{\textit{ATAC+Seq}}
& GET~\cite{fu2025get}                                  & 0.826 & 0.817 & 0.623 & 0.602 & 0.589 & 0.178 & 0.902 & 0.884 & 0.758 & 0.882 & 0.857 & 0.700 \\
& EpiBERT~\cite{javed2025epibert}                              & \underline{0.855} & \underline{0.833} & \underline{0.699} & 0.855 & \underline{0.849} & \underline{0.702} & 0.813 & 0.806 & 0.640 & 0.881 & \underline{0.873} & \underline{0.730} \\
\cmidrule(lr){1-14}
\rowcolor{mygray}
& \method{}                            & \textbf{0.893} & \textbf{0.887} & \textbf{0.788} & \textbf{0.948} & \textbf{0.937} & \textbf{0.873} & \textbf{0.979} & \textbf{0.978} & \textbf{0.956} & \textbf{0.940} & \textbf{0.928} & \textbf{0.863} \\

\bottomrule
\end{tabular}
}
\vspace{-0.1cm}
\end{table*}

\subsection{Training Configuration}
\label{sec:training_config}
We pre-train on PBMC169K, a multi-batch human PBMC scATAC-seq benchmark \citep{de2024systematic,jiao2025chromfound}, excluding the \texttt{BIO\_ddseq\_1} and \texttt{VIB\_10xv1\_1} batches. The initial gene vocabulary contains 5,257 highly variable PBMC10K genes. AlphaGenome embeddings are frozen inputs (dimension 3072); the multi-scale encoder uses dimension 2048, 16 heads, one peak self-attention layer, and two cross-scale blocks. Peak MLM training uses Adam, a peak learning rate of $10^{-3}$, 100 warmup updates, and an effective batch size of 32 for five epochs, taking approximately 45 hours on four H200 GPUs (Appendix~\ref{app:pretrain_details}).

Memory-efficient exact attention permits larger inputs, although its arithmetic cost remains quadratic in peak count. Appendix~\ref{app:efficiency} reports memory scaling and an inference-time comparison.

\subsection{Gene Expression Prediction}
\label{sec:gex_pred}
Following the Borzoi split~\citep{linder2025borzoi}, we hold out one quarter of genomic regions, including 1,161 of the initial 5,257 genes. Genes and peaks in these regions are excluded from task training and introduced at test time. We also hold out selected cell types. Experiments use four paired scATAC--scRNA datasets: PBMC10K~\citep{zheng2017massively}, BMMC~\citep{luecken2021sandbox}, Openproblem~\citep{open-problems-multimodal}, and Bone~\citep{to2024multi}. Gene sets and splits are shared across methods (Appendix~\ref{app:gex_pred_details}).

We compare non-sequence models (CMAE, scMoGNN, ChromFound, EpiAgent), sequence-based predictors (Borzoi, AlphaGenome, scooby), and accessibility-plus-sequence models (GET, EpiBERT). Table~\ref{tab:cross_omics_prediction_main} reports PCC, CCC, and $R^2$ on per-cell-type pseudobulk profiles. Additional EpiAgent diagnostics and PCA+Ridge comparisons, single-cell evaluation, and five-seed BMMC results are provided in Appendix~\ref{app:additional_gex}. Fixed-vocabulary models are included only for seen sequences in the main table.

\paragraph{Generalization and complementary evaluation.}
\method{} attains the best or tied-best pseudobulk PCC in all seen-sequence settings and six of eight unseen-sequence settings. It exceeds scooby in PCC throughout, but not in every metric: scooby has higher unseen-sequence CCC on Bone. At single-cell resolution, \method{} obtains unseen-sequence PCCs of 0.436, 0.382, 0.535, and 0.410 across the four datasets, while remaining competitive rather than uniformly best on seen sequences. PCA+Ridge is strong on seen genes (PCC 0.900 on PBMC10K and 0.972 on BMMC), emphasizing the value of testing beyond the training gene vocabulary.

\subsection{Cross-scale Task: Enhancer--Gene Linking}
\label{sec:peak_gene_enhancer_task}
We score peak--gene pairs by embedding cosine similarity and take the maximum over peaks overlapping a candidate enhancer. Ground truth comes from GeneHancer~\citep{fishilevich2017genehancer}, using eQTL or Capture Hi-C (CHi-C) evidence. We remove elements overlapping the target TSS and sample chromosome- and distance-matched negatives, excluding all GeneHancer-annotated links from the negative pool. The AUROC benchmark uses a balanced 1:1 positive-to-negative ratio.

\begin{figure}[t]
    \centering
    \includegraphics[width=\textwidth]{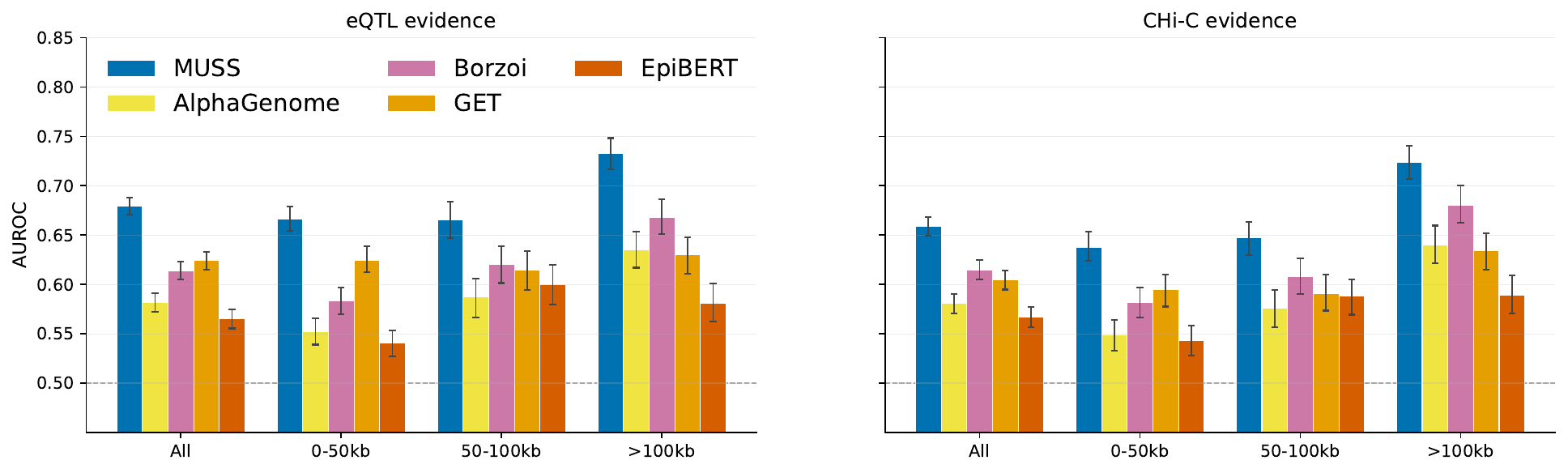}
    \caption{Distance-stratified AUROC on the balanced GeneHancer enhancer--gene linking benchmark. Left: eQTL-supported pairs; right: CHi-C-supported pairs. Error bars are 95\% bootstrap confidence intervals over 1,000 resamples; the dashed line denotes chance.}
    \label{fig:genehancer_main_auroc}
\end{figure}

Under a shared peak vocabulary and cosine-similarity protocol, \method{} outperforms AlphaGenome, Borzoi, GET, and EpiBERT across all distance bins and both evidence tiers (Figure~\ref{fig:genehancer_main_auroc}). We additionally evaluate per-gene retrieval over a $\pm1$ Mb cis window at approximately 1:36 class imbalance. \method{} achieves micro-AUPRC 0.1271, compared with 0.0834 for Borzoi and 0.0532 for PCA+Ridge, with especially large gains over the linear baseline beyond 100 kb. Full precision--recall and ranking results, including distance-stratified exceptions, are in Appendix~\ref{app:retrieval}.

\begin{table*}[htbp]

\centering
\caption{Cell clustering results on PBMC169K~\cite{de2024systematic} BIO\_ddseq\_1. Higher is better for all metrics. Values are mean $\pm$ standard deviation over 10 clustering seeds.}
\label{tab:cell_clustering_pbmc169k}
\footnotesize
\setlength{\tabcolsep}{4.2pt}
\renewcommand{\arraystretch}{1.08}
\begin{tabular}{ll|cccc}
\toprule
\textbf{Model type} & \textbf{Model} & ARI$\uparrow$ & FMI$\uparrow$ & NMI$\uparrow$ & AMI$\uparrow$ \\
\midrule

\multirow{4}{*}{\textit{Bulk}}
& AlphaGenome~\cite{avsec2026alphagenome}        & 0.1032$\pm$0.0000 & 0.3814$\pm$0.0000 & 0.1646$\pm$0.0000 & 0.1633$\pm$0.0000 \\
& Borzoi~\cite{linder2025borzoi}                 & 0.0837$\pm$0.0010 & 0.3577$\pm$0.0013 & 0.1450$\pm$0.0006 & 0.1437$\pm$0.0006 \\
& EpiBERT~\cite{javed2025epibert}                & 0.0909$\pm$0.0002 & 0.3642$\pm$0.0002 & 0.1499$\pm$0.0008 & 0.1486$\pm$0.0008 \\
& GET~\cite{fu2025get}                           & 0.0916$\pm$0.0002 & 0.3671$\pm$0.0004 & 0.1486$\pm$0.0001 & 0.1473$\pm$0.0001 \\

\cmidrule(lr){1-6}

\multirow{9}{*}{\textit{Single cell}}
& PeakVI~\cite{ashuach2022peakvi}                & 0.2915$\pm$0.0005 & 0.5041$\pm$0.0004 & 0.4623$\pm$0.0004 & 0.4615$\pm$0.0005 \\
& PoissonVI~\cite{martens2024poissonvi}          & 0.4103$\pm$0.0045 & 0.5994$\pm$0.0028 & 0.5656$\pm$0.0035 & 0.5649$\pm$0.0035 \\
& SnapATAC2~\cite{zhang2024snapatac2}            & 0.4126$\pm$0.0318 & 0.6149$\pm$0.0149 & 0.5761$\pm$0.0071 & 0.5754$\pm$0.0071 \\
& Signac~\cite{stuart2021signac}                 & 0.1885$\pm$0.0097 & 0.5249$\pm$0.0056 & 0.2852$\pm$0.0040 & 0.2839$\pm$0.0040 \\
& SCALE~\cite{xiong2019scale}                    & 0.4359$\pm$0.0181 & 0.6250$\pm$0.0101 & 0.5594$\pm$0.0045 & 0.5587$\pm$0.0045 \\
& SCALEX~\cite{xiong2022scalex}                  & 0.4087$\pm$0.0197 & 0.6063$\pm$0.0106 & 0.5519$\pm$0.0066 & 0.5512$\pm$0.0066 \\
& CASTLE~\cite{cui2024castle}                    & 0.3504$\pm$0.0042 & 0.5555$\pm$0.0026 & 0.5512$\pm$0.0013 & 0.5506$\pm$0.0013 \\
& scBasset~\cite{yuan2022scbasset}               & 0.4362$\pm$0.0012 & 0.6223$\pm$0.0008 & 0.5577$\pm$0.0008 & 0.5571$\pm$0.0008 \\
& ChromFound~\cite{jiao2025chromfound}           & \underline{0.4835$\pm$0.0082} & \underline{0.6604$\pm$0.0042} & \underline{0.5950$\pm$0.0036} & \underline{0.5944$\pm$0.0036} \\

\cmidrule(lr){1-6}

\rowcolor{mygray}
& {\method}                                      & \textbf{0.5377$\pm$0.0008} & \textbf{0.6973$\pm$0.0006} & \textbf{0.6085$\pm$0.0003} & \textbf{0.6079$\pm$0.0003} \\

\bottomrule
\end{tabular}
\end{table*}

\subsection{Cell-level Task: Zero-shot Cell Clustering}
\label{sec:cell_level_task}
We evaluate the held-out PBMC169K batch \texttt{BIO\_ddseq\_1} using the ChromFound protocol~\citep{jiao2025chromfound} (Appendix~\ref{app:cell_clustering_details}). \method{} leads all four metrics (Table~\ref{tab:cell_clustering_pbmc169k}), with ARI 0.5377 compared with 0.4835 for ChromFound and 0.1032 for AlphaGenome. The difference from the underlying sequence encoder indicates the importance of cell-contextualized representations. A PCA baseline (Appendix~\ref{app:pca_clustering}) and peak-MLM ablations (Appendix~\ref{app:additional_ablations}) further assess this contribution.

\section{Analysis Experiments}
\label{sec:analytical_study}

\subsection{Ablation Study}
\label{sec:ablation}

We conduct ablations on the BMMC dataset over three aspects: architecture, positional encoding, and training strategy (Table~\ref{tab:ablation_main}).

\paragraph{Sequence, interactions, and positional context.}
Random sequence embeddings markedly reduce unseen-sequence accuracy, and removing cross-scale interactions causes a still larger drop. Ablating distance reweighting, peak RoPE, or RBF-based distance bias also reduces performance; the RBF bias has the largest positional effect. These results support combining sequence content with explicit peak--gene interactions rather than relying on either alone.

\paragraph{Alternative sequence encoders.}
We test whether \method{} depends on AlphaGenome by replacing its frozen sequence features with Borzoi~\citep{linder2025borzoi} or NT-v3~\citep{ntv3_2025} across PBMC10K, BMMC, and OP (Table~\ref{tab:alternative_encoders}). Across all three datasets, the sequence-to-function foundation models AlphaGenome and Borzoi consistently outperform NT-v3 on unseen-sequence prediction, while differences on seen sequences are generally smaller. Their relative strengths vary across datasets: Borzoi achieves slightly higher unseen PCC and $R^2$ on PBMC10K, whereas AlphaGenome performs best across all unseen-sequence metrics on BMMC and OP. Despite these differences, AlphaGenome and Borzoi achieve broadly comparable performance, demonstrating that sequence-to-function foundation models provide strong sequence representations for \method{} and that its effectiveness does not depend on embeddings from any single pretrained encoder.

\begin{table*}[t]

\centering
\setlength{\tabcolsep}{2.5pt}
\renewcommand{\arraystretch}{1.12}
\caption{BMMC gene expression ablations. Differences from the full \method{} model are shown in parentheses.}
\label{tab:ablation_main}

\resizebox{\textwidth}{!}{
\begin{tabular}{ll|ccc|ccc}
\toprule
\textbf{Type} & \textbf{Ablation}
& \multicolumn{3}{c|}{\cellcolor{unseenpeach}\textbf{Unseen sequence}}
& \multicolumn{3}{c}{\cellcolor{seenblue}\textbf{Seen sequence}} \\
\cmidrule(lr){3-5}\cmidrule(lr){6-8}

\multicolumn{2}{c|}{}
& PCC$\uparrow$ & CCC$\uparrow$ & $R^2$$\uparrow$
& PCC$\uparrow$ & CCC$\uparrow$ & $R^2$$\uparrow$ \\
\midrule

\multirow{2}{*}{Arch}

& w/o Sequence embedding
& 0.196{\footnotesize\,(-0.549)}
& 0.132{\footnotesize\,(-0.600)}
& -0.330{\footnotesize\,(-0.837)}
& 0.958{\footnotesize\,(-0.011)}
& 0.947{\footnotesize\,(-0.016)}
& 0.890{\footnotesize\,(-0.036)} \\

& w/o Cross-scale Interact
& 0.009{\footnotesize\,(-0.736)}
& 0.002{\footnotesize\,(-0.730)}
& -1.422{\footnotesize\,(-1.929)}
& \underline{0.967}{\footnotesize\,(-0.002)}
& \underline{0.959}{\footnotesize\,(-0.004)}
& \underline{0.920}{\footnotesize\,(-0.006)} \\

\cmidrule(lr){1-8}

\multirow{3}{*}{Pos. Info}
& w/o Distance Reweighting
& \underline{0.736}{\footnotesize\,(-0.009)}
& \underline{0.722}{\footnotesize\,(-0.010)}
& \underline{0.480}{\footnotesize\,(-0.027)}
& \underline{0.967}{\footnotesize\,(-0.002)}
& 0.958{\footnotesize\,(-0.005)}
& 0.915{\footnotesize\,(-0.011)} \\

& w/o Peak RoPE
& 0.725{\footnotesize\,(-0.020)}
& 0.717{\footnotesize\,(-0.015)}
& 0.454{\footnotesize\,(-0.053)}
& 0.959{\footnotesize\,(-0.010)}
& 0.953{\footnotesize\,(-0.010)}
& 0.905{\footnotesize\,(-0.021)} \\

& w/o RBF distance
& 0.707{\footnotesize\,(-0.038)}
& 0.700{\footnotesize\,(-0.032)}
& 0.426{\footnotesize\,(-0.081)}
& 0.959{\footnotesize\,(-0.010)}
& 0.953{\footnotesize\,(-0.010)}
& 0.904{\footnotesize\,(-0.022)} \\

\cmidrule(lr){1-8}

Training
& w/o Peak MLM training
& 0.734{\footnotesize\,(-0.011)}
& 0.718{\footnotesize\,(-0.014)}
& 0.459{\footnotesize\,(-0.048)}
& 0.959{\footnotesize\,(-0.010)}
& 0.945{\footnotesize\,(-0.018)}
& 0.887{\footnotesize\,(-0.039)} \\

\midrule
\rowcolor{mygray}
\multicolumn{2}{l|}{\method{}}
& \textbf{0.745}
& \textbf{0.732}
& \textbf{0.507}
& \textbf{0.969}
& \textbf{0.963}
& \textbf{0.926} \\

\bottomrule
\end{tabular}
}
\end{table*}

\paragraph{Two-sided genomic context.}
An upstream-only ablation further supports the benefit of using peaks on both sides of a gene for seen-sequence prediction, while revealing mixed effects on unseen-sequence performance (Appendix~\ref{app:two_sided_context}).

\paragraph{Peak MLM transfers beyond expression prediction.}
Removing peak MLM weakens expression prediction (Table~\ref{tab:ablation_main}) and causes more pronounced losses in zero-shot cell clustering and enhancer--gene linking, including retrieval. These cross-task trends support learning transferable cell-level accessibility context beyond frozen sequence embeddings (Appendix~\ref{app:peak_mlm_transfer}).

\begin{table*}[t]
\centering
\setlength{\tabcolsep}{14.0pt}
\renewcommand{\arraystretch}{1.10}
\caption{Gene-expression prediction with different frozen DNA sequence encoders on PBMC10K, BMMC, and Openproblem (OP). Bold marks the best result per metric within each dataset and split, including ties.}
\label{tab:alternative_encoders}
\resizebox{\textwidth}{!}{%
\begin{tabular}{ll|ccc|ccc}
\toprule
\textbf{Dataset} & \textbf{Sequence Encoder}
& \multicolumn{3}{c|}{\cellcolor{unseenpeach}\textbf{Unseen sequence}}
& \multicolumn{3}{c}{\cellcolor{seenblue}\textbf{Seen sequence}} \\
\cmidrule(lr){3-5}\cmidrule(lr){6-8}
\multicolumn{2}{c|}{}
& PCC$\uparrow$ & CCC$\uparrow$ & $R^2$$\uparrow$
& PCC$\uparrow$ & CCC$\uparrow$ & $R^2$$\uparrow$ \\
\midrule
\multirow{3}{*}{PBMC10K}
& Borzoi      & \textbf{0.698} & 0.655          & \textbf{0.439} & \textbf{0.914} & 0.885          & 0.798 \\
& NT-v3       & 0.459          & 0.409          & 0.083          & 0.912          & 0.887          & 0.801 \\
\rowcolor{mygray}
& AlphaGenome & 0.684          & \textbf{0.663} & 0.409          & 0.909          & \textbf{0.895} & \textbf{0.811} \\
\cmidrule(lr){1-8}
\multirow{3}{*}{BMMC}
& Borzoi      & 0.695          & 0.692          & 0.430          & \textbf{0.986} & \textbf{0.984} & \textbf{0.969} \\
& NT-v3       & 0.452          & 0.437          & 0.037          & 0.983          & 0.980          & 0.961 \\
\rowcolor{mygray}
& AlphaGenome & \textbf{0.745} & \textbf{0.732} & \textbf{0.507} & 0.969          & 0.963          & 0.926 \\
\cmidrule(lr){1-8}
\multirow{3}{*}{OP}
& Borzoi      & 0.707          & 0.696          & 0.474          & \textbf{0.996} & \textbf{0.996} & \textbf{0.991} \\
& NT-v3       & 0.549          & 0.512          & 0.249          & \textbf{0.996} & 0.995          & 0.990 \\
\rowcolor{mygray}
& AlphaGenome & \textbf{0.755} & \textbf{0.734} & \textbf{0.552} & 0.995          & 0.995          & 0.990 \\
\bottomrule
\end{tabular}%
}
\end{table*}

\begin{figure*}[h]
    \centering

    \begin{minipage}[t]{0.49\textwidth}
        \centering
        \includegraphics[width=\linewidth]{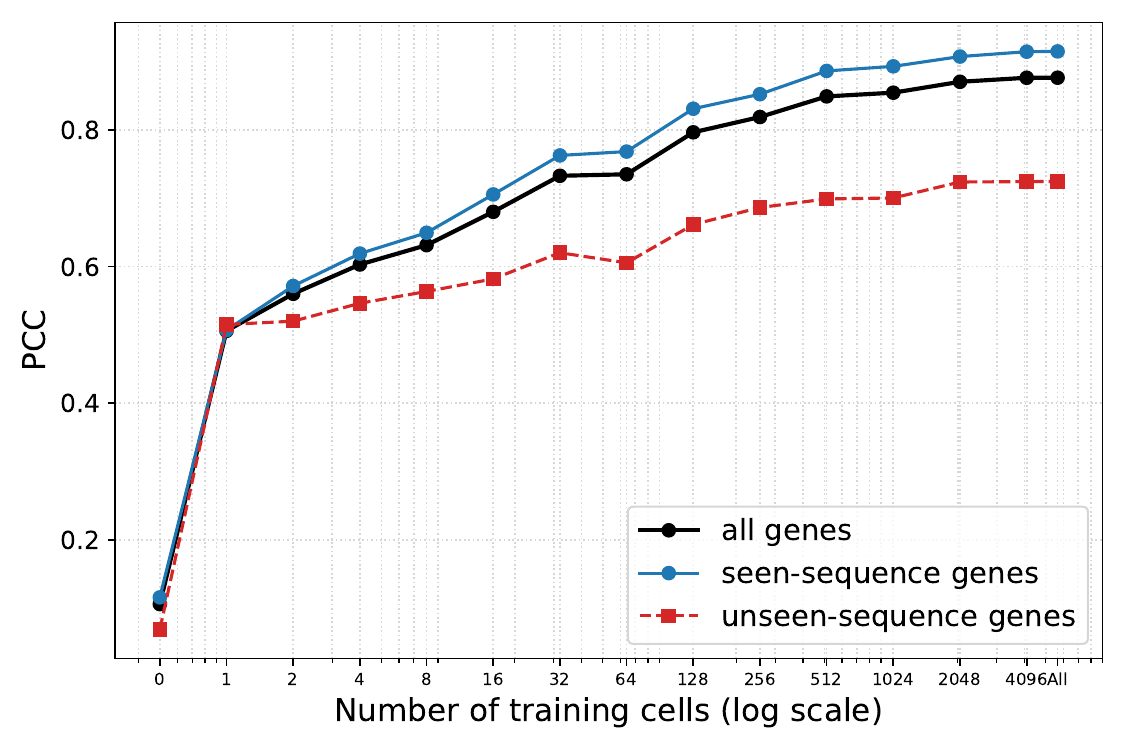}
    \end{minipage}
    \hfill
    \begin{minipage}[t]{0.49\textwidth}
        \centering
        \includegraphics[width=\linewidth]{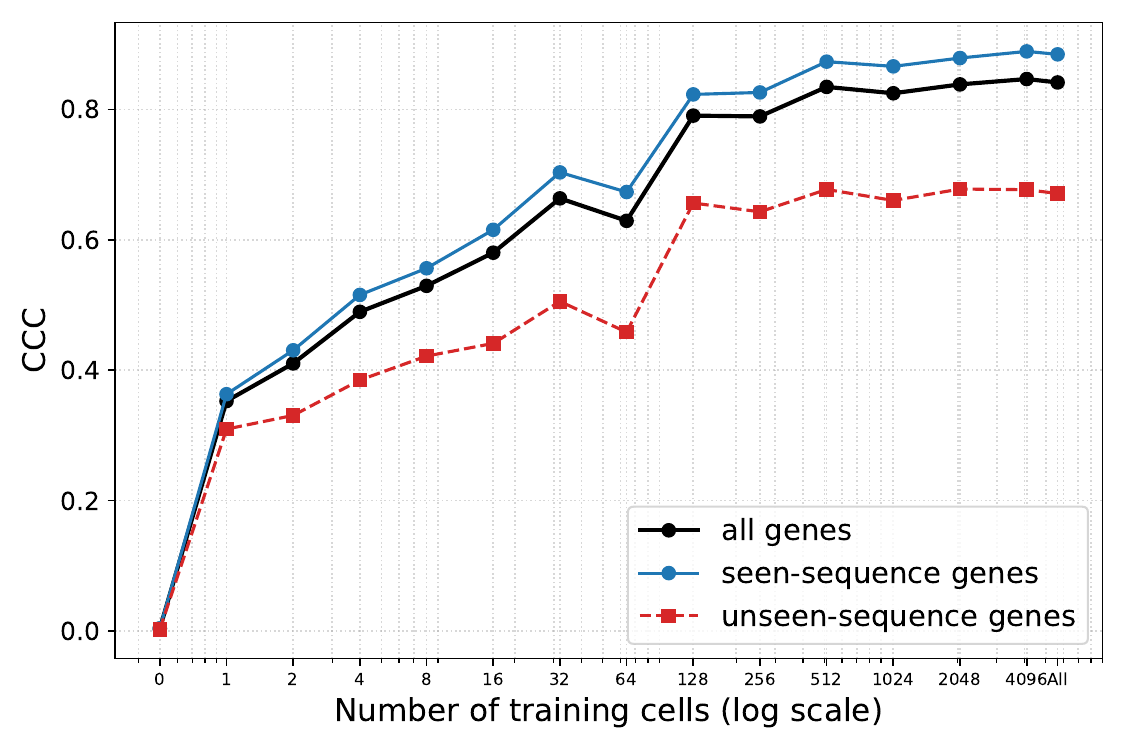}
    \end{minipage}

    \caption{Few-shot scaling curves for gene expression prediction on PBMC10K. The two panels show the PCC and CCC of \method{} as the number of paired cells used for fine-tuning increases.}
    \label{fig:fewshot_pbmc10k}

\end{figure*}

\subsection{Few-shot Gene Expression Prediction}
\label{sec:fewshot_gex}

We vary the number of paired PBMC10K training cells while initializing from the same peak-MLM-pretrained checkpoint. With one paired cell, PCC and CCC are approximately 0.5; performance improves with additional cells and begins to saturate after a few hundred (Figure~\ref{fig:fewshot_pbmc10k}). Both seen and unseen genes benefit from additional supervision. This demonstrates data-efficient \emph{adaptation}, not elimination of the data required for pretraining. Appendix~\ref{app:pbmc10k_fewshot} specifies the subsets and optimization budgets.

\subsection{Motif-guided Regulatory Prioritization}
\label{sec:biological_application}
We assess whether sequence-grounded embeddings prioritize candidate TF--peak--gene relationships beyond genomic proximity. A motif-informed benchmark combines open peaks with CollecTRI~\citep{muller2023collectri} (43,175 TF--gene interactions across 1,186 TFs). Motifs are detected with FIMO~\citep{grant2011fimo,bailey2015meme,fornes2020jaspar}. A broad positive requires at least one motif for a known regulator of the gene; the conservative definition requires motifs for at least two distinct relevant TFs and excludes SP/KLF-only positives. Controls are matched by gene and distance, lack motifs for that gene's regulators, and retain motifs for other TFs. Candidate peaks are ranked by peak--gene embedding cosine similarity (Appendix~\ref{app:bio_app}).

 \begin{figure}
     \centering
     \includegraphics[width=1\linewidth]{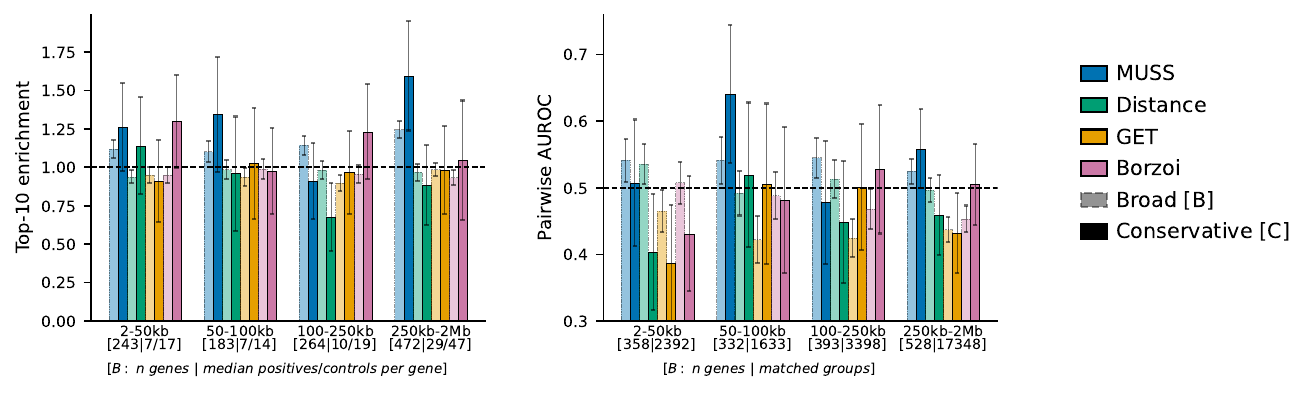}
     \caption{DNA sequence motif–informed peak–gene prioritization across genomic distances. Left: top-10 enrichment of known TF–gene links among top-ranked peaks (averaged per gene). Right: pairwise AUROC comparing positive peaks (broad and conservative definitions) to distance-matched controls. Error bars show 95\% CI; dashed line indicates chance.}
     \label{fig:motif_prioritisation}
 \end{figure}

In Figure~\ref{fig:motif_prioritisation}, \method{} exceeds the distance baseline across the broad setting, with the largest gains at 250 kb--2 Mb. Under the conservative definition, pairwise AUROC at 50--100 kb is $0.641\pm0.103$, compared with $0.519\pm0.109$ for distance alone. These results support regulatory prioritization beyond proximity. They remain hypothesis-generating: motif presence does not establish TF binding or causality, and differences in how baselines encode genomic position affect this comparison (Appendix~\ref{app:bio_app}). Candidate links could guide subsequent CRISPR perturbation or enhancer engineering experiments~\citep{fulco2019activity,gasperini2019genomewide,melnikov2012systematic,kinney2014massively,king2020synthetic,deboer2020deciphering}.

\section{Conclusion, Limitations, and Broader Impact}
\label{sec:conclusion}
\method{} integrates DNA-derived peak and gene representations with cell-specific accessibility through multi-scale attention. The expression, regulatory-linking, and clustering experiments support transferable representations, including expression prediction beyond the task-training vocabulary. Additional encoder and pretraining ablations show that both sequence information and learned accessibility context matter.

\paragraph{Limitations.}
\method{} depends on a substantial frozen sequence encoder and paired multi-omic data for expression training. Alternative encoders remain useful but differ in unseen-sequence accuracy. Exact peak self-attention has quadratic arithmetic cost, even with memory-efficient kernels; sparse or low-rank alternatives are future work. Inference is slower than lightweight baselines (Appendix~\ref{app:efficiency}). Enhancer--gene evaluation is restricted to the training peak and gene vocabularies and does not establish unseen-link generalization. The upstream-only ablation does not isolate the two directions of cross-scale information flow. No single method dominates every metric or evaluation resolution.

\paragraph{Broader impact.}
The model may help prioritize regulatory hypotheses and experiments, but association scores are not causal or clinical evidence. Unvalidated use in therapeutic targeting or sequence engineering could produce misleading conclusions or unintended biological effects. Experimental validation and appropriate oversight remain necessary.

\label{main:lastpage}
\clearpage

\begin{ack}
\textbf{Funding.}
This project was funded by Open Targets under OTAR 3090,  Drug2Cell (D2C): developing foundation models for scRNA-seq data.

\textbf{Acknowledgments.}
We thank all members of the Lotfollahi Group for their valuable feedback and suggestions during internal discussions.

\textbf{Competing interests.}
The authors declare no competing interests.
\end{ack}

\bibliographystyle{plainnat}
\bibliography{references}

\begin{thebibliography}{50}
\providecommand{\natexlab}[1]{#1}
\providecommand{\url}[1]{\texttt{#1}}
\expandafter\ifx\csname urlstyle\endcsname\relax
  \providecommand{\doi}[1]{doi: #1}\else
  \providecommand{\doi}{doi: \begingroup \urlstyle{rm}\Url}\fi

\bibitem[Ashuach et~al.(2022)Ashuach, Reidenbach, Gayoso, and Yosef]{ashuach2022peakvi}
Tal Ashuach, Daniel~A. Reidenbach, Adam Gayoso, and Nir Yosef.
\newblock {PeakVI}: A deep generative model for single-cell chromatin accessibility analysis.
\newblock \emph{Cell Reports Methods}, 2\penalty0 (3):\penalty0 100182, 2022.
\newblock \doi{10.1016/j.crmeth.2022.100182}.

\bibitem[Avsec et~al.(2026)Avsec, Latysheva, Cheng, Novati, Taylor, Ward, Bycroft, Nicolaisen, Arvaniti, Pan, Thomas, Dutordoir, Perino, De, Karollus, Gayoso, Sargeant, Mottram, Wong, Drotar, Kosiorek, Senior, Tanburn, Applebaum, Basu, Hassabis, and Kohli]{avsec2026alphagenome}
{\v{Z}}iga Avsec, Natasha Latysheva, Jun Cheng, Guido Novati, Kyle~R. Taylor, Tom Ward, Clare Bycroft, Lauren Nicolaisen, Eirini Arvaniti, Joshua Pan, Raina Thomas, Vincent Dutordoir, Matteo Perino, Soham De, Alexander Karollus, Adam Gayoso, Toby Sargeant, Anne Mottram, Lai~Hong Wong, Pavol Drotar, Adam Kosiorek, Andrew Senior, Richard Tanburn, Taylor Applebaum, Souradeep Basu, Demis Hassabis, and Pushmeet Kohli.
\newblock Advancing regulatory variant effect prediction with {AlphaGenome}.
\newblock \emph{Nature}, 649\penalty0 (8099):\penalty0 1206--1218, 2026.
\newblock \doi{10.1038/s41586-025-10014-0}.

\bibitem[Baek et~al.(2017)Baek, Goldstein, and Hager]{baek2017bivariate}
Seungsoo Baek, Isaac Goldstein, and Gordon~L Hager.
\newblock Bivariate genomic footprinting detects changes in transcription factor activity.
\newblock \emph{Cell Reports}, 19\penalty0 (8):\penalty0 1710--1722, 2017.

\bibitem[Bailey et~al.(2015)Bailey, Johnson, Grant, and Noble]{bailey2015meme}
Timothy~L Bailey, James Johnson, Charles~E Grant, and William~S Noble.
\newblock The meme suite.
\newblock \emph{Nucleic Acids Research}, 43\penalty0 (W1):\penalty0 W39--W49, 2015.

\bibitem[Boshar et~al.(2025)Boshar, Evans, Tang, Picard, Adel, Lorbeer, Rajesh, Karch, Sidbon, Emms, Mendoza-Revilla, Al-Ani, Seitz, Schiff, Bornachot, Hernandez, Lopez, Laterre, Beguir, Koo, Kuleshov, Stark, de~Almeida, and Pierrot]{ntv3_2025}
Sam Boshar, Benjamin Evans, Ziqi Tang, Armand Picard, Yanis Adel, Franziska~K. Lorbeer, Chandana Rajesh, Tristan Karch, Shawn Sidbon, David Emms, Javier Mendoza-Revilla, Fatimah Al-Ani, Evan Seitz, Yair Schiff, Yohan Bornachot, Ariana Hernandez, Marie Lopez, Alexandre Laterre, Karim Beguir, Peter Koo, Volodymyr Kuleshov, Alexander Stark, Bernardo~P. de~Almeida, and Thomas Pierrot.
\newblock A foundational model for joint sequence-function multi-species modeling at scale for long-range genomic prediction.
\newblock \emph{bioRxiv}, 2025.
\newblock \doi{10.64898/2025.12.22.695963}.

\bibitem[Brixi et~al.(2026)Brixi, Durrant, Ku, Naghipourfar, Poli, Sun, Brockman, Chang, Fanton, Gonzalez, et~al.]{brixi2025evo2}
Garyk Brixi, Matthew~G Durrant, Jerome Ku, Mohsen Naghipourfar, Michael Poli, Gwanggyu Sun, Greg Brockman, Daniel Chang, Alison Fanton, Gabriel~A Gonzalez, et~al.
\newblock Genome modelling and design across all domains of life with evo 2.
\newblock \emph{Nature}, pages 1--13, 2026.

\bibitem[Burkhardt et~al.(2022)Burkhardt, Luecken, Benz, Holderrieth, Bloom, Lance, Chow, and Holbrook]{open-problems-multimodal}
Daniel Burkhardt, Malte Luecken, Andrew Benz, Peter Holderrieth, Jonathan Bloom, Christopher Lance, Ashley Chow, and Ryan Holbrook.
\newblock Open problems -- multimodal single-cell integration.
\newblock \url{https://kaggle.com/competitions/open-problems-multimodal}, 2022.
\newblock Kaggle.

\bibitem[Chen et~al.(2025)Chen, Li, Cui, Wang, Jiang, Lin, Li, Gao, Lv, and Jiang]{chen2025epiagent}
Xiaoyang Chen, Keyi Li, Xuejian Cui, Zian Wang, Qun Jiang, Jiacheng Lin, Zhen Li, Zijing Gao, Hairong Lv, and Rui Jiang.
\newblock {EpiAgent}: Foundation model for single-cell epigenomics.
\newblock \emph{Nature Methods}, 22\penalty0 (11):\penalty0 2316--2327, 2025.
\newblock \doi{10.1038/s41592-025-02822-z}.

\bibitem[Cui et~al.(2024{\natexlab{a}})Cui, Wang, Maan, Pang, Luo, Duan, and Wang]{cui2024scgpt}
Haotian Cui, Chloe Wang, Hassaan Maan, Kuan Pang, Fengning Luo, Nan Duan, and Bo~Wang.
\newblock {scGPT}: Toward building a foundation model for single-cell multi-omics using generative {AI}.
\newblock \emph{Nature Methods}, 2024{\natexlab{a}}.
\newblock \doi{10.1038/s41592-024-02201-0}.

\bibitem[Cui et~al.(2024{\natexlab{b}})Cui, Chen, Li, Gao, Chen, and Jiang]{cui2024castle}
Xuejian Cui, Xiaoyang Chen, Zhen Li, Zijing Gao, Shengquan Chen, and Rui Jiang.
\newblock Discrete latent embedding of single-cell chromatin accessibility sequencing data for uncovering cell heterogeneity.
\newblock \emph{Nature Computational Science}, 4:\penalty0 346--359, 2024{\natexlab{b}}.
\newblock \doi{10.1038/s43588-024-00625-4}.

\bibitem[Dalla-Torre et~al.(2025)Dalla-Torre, Gonzalez, Mendoza-Revilla, Lopez~Carranza, Grzywaczewski, Oteri, Dallago, Trop, De~Almeida, Sirelkhatim, et~al.]{dalla-torre2024nucleotide}
Hugo Dalla-Torre, Liam Gonzalez, Javier Mendoza-Revilla, Nicolas Lopez~Carranza, Adam~Henryk Grzywaczewski, Francesco Oteri, Christian Dallago, Evan Trop, Bernardo~P De~Almeida, Hassan Sirelkhatim, et~al.
\newblock Nucleotide transformer: building and evaluating robust foundation models for human genomics.
\newblock \emph{Nature Methods}, 22\penalty0 (2):\penalty0 287--297, 2025.

\bibitem[de~Boer et~al.(2020)de~Boer, Vaishnav, Sadeh, Abeyta, Friedman, and Regev]{deboer2020deciphering}
Carl~G de~Boer, Eeshit~Dhaval Vaishnav, Ronen Sadeh, Esteban~Luis Abeyta, Nir Friedman, and Aviv Regev.
\newblock Deciphering eukaryotic gene-regulatory logic with 100 million random promoters.
\newblock \emph{Nature biotechnology}, 38\penalty0 (1):\penalty0 56--65, 2020.

\bibitem[De~Rop et~al.(2024)De~Rop, Hulselmans, Flerin, Soler-Vila, Rafels, Christiaens, Gonzalez-Blas, Marchese, Caratu, Poovathingal, et~al.]{de2024systematic}
Florian~V De~Rop, Gert Hulselmans, Chris Flerin, Paula Soler-Vila, Albert Rafels, Valerie Christiaens, Carmen~Bravo Gonzalez-Blas, Domenica Marchese, Ginevra Caratu, Suresh Poovathingal, et~al.
\newblock Systematic benchmarking of single-cell atac-sequencing protocols.
\newblock \emph{Nature biotechnology}, 42\penalty0 (6):\penalty0 916--926, 2024.

\bibitem[Fishilevich et~al.(2017)Fishilevich, Nudel, Rappaport, Hadar, Plaschkes, Iny~Stein, Rosen, Kohn, Twik, Safran, et~al.]{fishilevich2017genehancer}
Simon Fishilevich, Ron Nudel, Noa Rappaport, Rotem Hadar, Inbar Plaschkes, Tsippi Iny~Stein, Naomi Rosen, Asher Kohn, Michal Twik, Marilyn Safran, et~al.
\newblock {GeneHancer}: Genome-wide integration of enhancers and target genes in {GeneCards}.
\newblock \emph{Database}, 2017:\penalty0 bax028, 2017.

\bibitem[Fornes et~al.(2020)Fornes, Castro-Mondragon, Khan, van~der Lee, Zhang, Richmond, Modi, Correard, Gheorghe, Baranasic, et~al.]{fornes2020jaspar}
Oriol Fornes, Jaime~A Castro-Mondragon, Aziz Khan, Robin van~der Lee, Xi~Zhang, Phillip~A Richmond, Bhavi~P Modi, Sylvain Correard, Mariana Gheorghe, Dario Baranasic, et~al.
\newblock Jaspar 2020: update of the open-access database of transcription factor binding profiles.
\newblock \emph{Nucleic Acids Research}, 48\penalty0 (D1):\penalty0 D87--D92, 2020.

\bibitem[Fu et~al.(2025)Fu, Mo, Buendia, Laurent, Shao, Alvarez-Torres, Yu, Tan, Su, Sagatelian, Ferrando, Ciccia, Lan, Owens, Palomero, Xing, and Rabadan]{fu2025get}
Xi~Fu, Shentong Mo, Alejandro Buendia, Anouchka~P. Laurent, Anqi Shao, Maria del~Mar Alvarez-Torres, Tianji Yu, Jimin Tan, Jiayu Su, Romella Sagatelian, Adolfo~A. Ferrando, Alberto Ciccia, Yanyan Lan, David~M. Owens, Teresa Palomero, Eric~P. Xing, and Raul Rabadan.
\newblock A foundation model of transcription across human cell types.
\newblock \emph{Nature}, 637\penalty0 (8047):\penalty0 965--973, 2025.
\newblock \doi{10.1038/s41586-024-08391-z}.

\bibitem[Fulco et~al.(2019)Fulco, Nasser, Jones, Munson, Bergman, Subramanian, Grossman, Anyoha, Doughty, Patwardhan, et~al.]{fulco2019activity}
Charles~P Fulco, Joseph Nasser, Thouis~R Jones, Glen Munson, Drew~T Bergman, Vidya Subramanian, Sharon~R Grossman, Rockwell Anyoha, Benjamin~R Doughty, Tejal~A Patwardhan, et~al.
\newblock Activity-by-contact model of enhancer--promoter regulation from thousands of crispr perturbations.
\newblock \emph{Nature genetics}, 51\penalty0 (12):\penalty0 1664--1669, 2019.

\bibitem[Gasperini et~al.(2019)Gasperini, Hill, McFaline-Figueroa, Martin, Kim, Zhang, Jackson, Leith, Schreiber, Noble, et~al.]{gasperini2019genomewide}
Molly Gasperini, Andrew~J Hill, Jose~L McFaline-Figueroa, Brian Martin, Stephanie Kim, Michelle~D Zhang, Devin Jackson, Adrian Leith, Jennifer Schreiber, William~S Noble, et~al.
\newblock A genome-wide framework for mapping gene regulation via cellular genetic screens.
\newblock \emph{Cell}, 176\penalty0 (1-2):\penalty0 377--390.e19, 2019.
\newblock \doi{10.1016/j.cell.2018.11.029}.

\bibitem[Grant et~al.(2011)Grant, Bailey, and Noble]{grant2011fimo}
Charles~E Grant, Timothy~L Bailey, and William~S Noble.
\newblock Fimo: scanning for occurrences of a given motif.
\newblock \emph{Bioinformatics}, 27\penalty0 (7):\penalty0 1017--1018, 2011.

\bibitem[Gschwind et~al.(2023)Gschwind, Mualim, Karbalayghareh, Sheth, Dey, Jagoda, Nurtdinov, Xi, Tan, Jones, et~al.]{gschwind2023encyclopedia}
Andreas~R. Gschwind, Kristy~S. Mualim, Alireza Karbalayghareh, Maya~U. Sheth, Kushal~K. Dey, Evelyn Jagoda, Ramil~N. Nurtdinov, Wang Xi, Anthony~S. Tan, Hank Jones, et~al.
\newblock An encyclopedia of enhancer-gene regulatory interactions in the human genome.
\newblock \emph{bioRxiv}, 2023.

\bibitem[Hao et~al.(2024)Hao, Gong, Zeng, Liu, Guo, Cheng, Wang, Ma, Zhang, and Song]{hao2024scfoundation}
Minsheng Hao, Jing Gong, Xin Zeng, Chiming Liu, Yucheng Guo, Xingyi Cheng, Taifeng Wang, Jianzhu Ma, Xuegong Zhang, and Le~Song.
\newblock Large-scale foundation model on single-cell transcriptomics.
\newblock \emph{Nature Methods}, 2024.
\newblock \doi{10.1038/s41592-024-02305-7}.

\bibitem[Hingerl et~al.(2025)Hingerl, Martens, Karollus, Manz, Buenrostro, Theis, and Gagneur]{hingerl2025scooby}
Johannes~C. Hingerl, Laura~D. Martens, Alexander Karollus, Trevor Manz, Jason~D. Buenrostro, Fabian~J. Theis, and Julien Gagneur.
\newblock Scooby: Modeling multimodal genomic profiles from {DNA} sequence at single-cell resolution.
\newblock \emph{Nature Methods}, 22\penalty0 (11):\penalty0 2275--2285, 2025.
\newblock \doi{10.1038/s41592-025-02854-5}.

\bibitem[Javed et~al.(2025)Javed, Weingarten, Sehanobish, Roberts, Dubey, Choromanski, and Bernstein]{javed2025epibert}
Nauman Javed, Thomas Weingarten, Arijit Sehanobish, Adam Roberts, Avinava Dubey, Krzysztof Choromanski, and Bradley~E. Bernstein.
\newblock A multi-modal transformer for cell type-agnostic regulatory predictions.
\newblock \emph{Cell Genomics}, 5\penalty0 (2), 2025.
\newblock \doi{10.1016/j.xgen.2025.100762}.
\newblock URL \url{https://www.cell.com/cell-genomics/fulltext/S2666-979X(25)00018-7}.

\bibitem[Jiao et~al.(2025)Jiao, Liu, Zhang, Guo, Wu, Jiang, Li, Zhang, Han, Gao, Qi, and Cheng]{jiao2025chromfound}
Yifeng Jiao, Yuchen Liu, Yu~Zhang, Xin Guo, Yushuai Wu, Chen Jiang, Jiyang Li, Hongwei Zhang, Limei Han, Xin Gao, Yuan Qi, and Yuan Cheng.
\newblock {ChromFound}: Towards a universal foundation model for single-cell chromatin accessibility data.
\newblock \emph{arXiv preprint arXiv:2505.12638}, 2025.

\bibitem[King et~al.(2020)King, Hong, Shepherdson, Granas, Maricque, and Cohen]{king2020synthetic}
Dana~M King, Clarice Kit~Yee Hong, James~L Shepherdson, David~M Granas, Brett~B Maricque, and Barak~A Cohen.
\newblock Synthetic and genomic regulatory elements reveal aspects of cis-regulatory grammar in mouse embryonic stem cells.
\newblock \emph{elife}, 9:\penalty0 e41279, 2020.

\bibitem[Kinney and McCandlish(2019)]{kinney2014massively}
Justin~B Kinney and David~M McCandlish.
\newblock Massively parallel assays and quantitative sequence--function relationships.
\newblock \emph{Annual review of genomics and human genetics}, 20\penalty0 (1):\penalty0 99--127, 2019.

\bibitem[Lambert et~al.(2018)Lambert, Jolma, Campitelli, Das, Yin, Albu, Chen, Taipale, Hughes, and Weirauch]{lambert2018human}
Samuel~A. Lambert, Arttu Jolma, Laura~F. Campitelli, Pratyush~K. Das, Yimeng Yin, Mihai Albu, Xiaoting Chen, Jussi Taipale, Timothy~R. Hughes, and Matthew~T. Weirauch.
\newblock The human transcription factors.
\newblock \emph{Cell}, 172\penalty0 (4):\penalty0 650--665, February 2018.

\bibitem[Linder et~al.(2025)Linder, Srivastava, Yuan, Agarwal, and Kelley]{linder2025borzoi}
Johannes Linder, Divyanshi Srivastava, Han Yuan, Vikram Agarwal, and David~R. Kelley.
\newblock Predicting {RNA}-seq coverage from {DNA} sequence as a unifying model of gene regulation.
\newblock \emph{Nature Genetics}, 57\penalty0 (4):\penalty0 949--961, 2025.
\newblock \doi{10.1038/s41588-024-02053-6}.

\bibitem[Luecken et~al.(2021)Luecken, Burkhardt, Cannoodt, Lance, Agrawal, Aliee, Chen, Deconinck, Detweiler, Granados, et~al.]{luecken2021sandbox}
Malte~D. Luecken, Daniel~Bernard Burkhardt, Robrecht Cannoodt, Christopher Lance, Aditi Agrawal, Hananeh Aliee, Ann~T. Chen, Louise Deconinck, Angela~M. Detweiler, Alejandro~A. Granados, et~al.
\newblock A sandbox for prediction and integration of {DNA}, {RNA}, and proteins in single cells.
\newblock In \emph{Thirty-Fifth Conference on Neural Information Processing Systems Datasets and Benchmarks Track (Round 2)}, 2021.

\bibitem[Martens et~al.(2024)Martens, Fischer, Y{\'e}pez, Theis, and Gagneur]{martens2024poissonvi}
Laura~D. Martens, David~S. Fischer, Vicente~A. Y{\'e}pez, Fabian~J. Theis, and Julien Gagneur.
\newblock Modeling fragment counts improves single-cell {ATAC}-seq analysis.
\newblock \emph{Nature Methods}, 21\penalty0 (1):\penalty0 28--31, 2024.
\newblock \doi{10.1038/s41592-023-02112-6}.

\bibitem[Melnikov et~al.(2012)Melnikov, Murugan, Zhang, Tesileanu, Wang, Rogov, Feizi, Gnirke, Callan, Kinney, et~al.]{melnikov2012systematic}
Alexandre Melnikov, Anand Murugan, Xiaolan Zhang, Tiberiu Tesileanu, Li~Wang, Peter Rogov, Soheil Feizi, Andreas Gnirke, Curtis~G Callan, Justin~B Kinney, et~al.
\newblock Systematic dissection and optimization of inducible enhancers in human cells using a massively parallel reporter assay.
\newblock \emph{Nature biotechnology}, 30\penalty0 (3):\penalty0 271--277, 2012.

\bibitem[M{\"u}ller-Dott et~al.(2023)M{\"u}ller-Dott, Tsirvouli, Vazquez, Ramirez~Flores, Badia-i Mompel, Fallegger, T{\"u}rei, L{\ae}greid, and Saez-Rodriguez]{muller2023collectri}
Sophia M{\"u}ller-Dott, Eirini Tsirvouli, Miguel Vazquez, Ricardo~O Ramirez~Flores, Pau Badia-i Mompel, Robin Fallegger, D{\'e}nes T{\"u}rei, Astrid L{\ae}greid, and Julio Saez-Rodriguez.
\newblock Expanding the coverage of regulons from high-confidence prior knowledge for accurate estimation of transcription factor activities.
\newblock \emph{Nucleic Acids Research}, 51\penalty0 (20):\penalty0 10934--10949, 2023.
\newblock \doi{10.1093/nar/gkad841}.

\bibitem[Nguyen et~al.(2023)Nguyen, Poli, Faizi, Thomas, Birch-Sykes, Wornow, Patel, Rabideau, Massaroli, Bengio, et~al.]{nguyen2023hyenadna}
Eric Nguyen, Michael Poli, Marjan Faizi, Armin Thomas, Callum Birch-Sykes, Michael Wornow, Aman Patel, Clayton Rabideau, Stefano Massaroli, Yoshua Bengio, et~al.
\newblock {HyenaDNA}: Long-range genomic sequence modeling at single nucleotide resolution.
\newblock In \emph{Advances in Neural Information Processing Systems (NeurIPS)}, 2023.
\newblock URL \url{https://openreview.net/pdf?id=ubzNoJjOKj}.

\bibitem[Schoenfelder and Fraser(2019)]{schoenfelder2019long}
Stefan Schoenfelder and Peter Fraser.
\newblock Long-range enhancer--promoter contacts in gene expression control.
\newblock \emph{Nature Reviews Genetics}, 20\penalty0 (8):\penalty0 437--455, 2019.

\bibitem[Scholkopf et~al.(1997)Scholkopf, Sung, Burges, Girosi, Niyogi, Poggio, and Vapnik]{scholkopf1997comparing}
Bernhard Scholkopf, Kah-Kay Sung, Christopher~JC Burges, Federico Girosi, Partha Niyogi, Tomaso Poggio, and Vladimir Vapnik.
\newblock Comparing support vector machines with gaussian kernels to radial basis function classifiers.
\newblock \emph{IEEE transactions on Signal Processing}, 45\penalty0 (11):\penalty0 2758--2765, 1997.

\bibitem[Stuart et~al.(2021)Stuart, Srivastava, Madad, Lareau, and Satija]{stuart2021signac}
Tim Stuart, Avi Srivastava, Shaista Madad, Caleb Lareau, and Rahul Satija.
\newblock Single-cell chromatin state analysis with {Signac}.
\newblock \emph{Nature Methods}, 18\penalty0 (11):\penalty0 1333--1341, 2021.
\newblock \doi{10.1038/s41592-021-01282-5}.

\bibitem[Su et~al.(2024)Su, Ahmed, Lu, Pan, Bo, and Liu]{su2024roformer}
Jianlin Su, Murtadha Ahmed, Yu~Lu, Shengfeng Pan, Wen Bo, and Yunfeng Liu.
\newblock Roformer: Enhanced transformer with rotary position embedding.
\newblock \emph{Neurocomputing}, 568:\penalty0 127063, 2024.

\bibitem[Theodoris et~al.(2023)Theodoris, Xiao, Chopra, Chaffin, Al~Sayed, Hill, Mantineo, Brydon, Zeng, Liu, and Ellinor]{theodoris2023geneformer}
Christina~V. Theodoris, Ling Xiao, Anant Chopra, Mark~D. Chaffin, Zeina~R. Al~Sayed, Matthew~C. Hill, Helene Mantineo, Elizabeth~M. Brydon, Zexian Zeng, X.~Shirley Liu, and Patrick~T. Ellinor.
\newblock Transfer learning enables predictions in network biology.
\newblock \emph{Nature}, 2023.
\newblock \doi{10.1038/s41586-023-06139-9}.

\bibitem[To et~al.(2024)To, Fei, Pett, Roberts, Blain, Pola{\'n}ski, Li, Yayon, He, Xu, et~al.]{to2024multi}
Ken To, Lijiang Fei, J.~Patrick Pett, Kenny Roberts, Raphael Blain, Krzysztof Pola{\'n}ski, Tong Li, Nadav Yayon, Peng He, Chuan Xu, et~al.
\newblock A multi-omic atlas of human embryonic skeletal development.
\newblock \emph{Nature}, 635\penalty0 (8039):\penalty0 657--667, 2024.

\bibitem[Vaswani et~al.(2017)Vaswani, Shazeer, Parmar, Uszkoreit, Jones, Gomez, Kaiser, and Polosukhin]{vaswani2017attention}
Ashish Vaswani, Noam Shazeer, Niki Parmar, Jakob Uszkoreit, Llion Jones, Aidan~N Gomez, {\L}ukasz Kaiser, and Illia Polosukhin.
\newblock Attention is all you need.
\newblock \emph{Advances in neural information processing systems}, 30, 2017.

\bibitem[Wen et~al.(2022)Wen, Ding, Jin, Wang, Xie, and Tang]{wen2022scmognn}
Hongzhi Wen, Jiayuan Ding, Wei Jin, Yiqi Wang, Yuying Xie, and Jiliang Tang.
\newblock Graph neural networks for multimodal single-cell data integration.
\newblock In \emph{Proceedings of the 28th ACM SIGKDD Conference on Knowledge Discovery and Data Mining}, pages 4153--4163, 2022.

\bibitem[Xiong et~al.(2019)Xiong, Xu, Tian, Shao, Tang, Gao, Zhang, Jiang, and Zhang]{xiong2019scale}
Lei Xiong, Kui Xu, Kang Tian, Yanqiu Shao, Lei Tang, Ge~Gao, Michael Zhang, Tao Jiang, and Qiangfeng~Cliff Zhang.
\newblock {SCALE} method for single-cell {ATAC}-seq analysis via latent feature extraction.
\newblock \emph{Nature Communications}, 10\penalty0 (1):\penalty0 4576, 2019.
\newblock \doi{10.1038/s41467-019-12630-7}.

\bibitem[Xiong et~al.(2022)Xiong, Tian, Li, Ning, Gao, and Zhang]{xiong2022scalex}
Lei Xiong, Kang Tian, Yuzhe Li, Weixi Ning, Xin Gao, and Qiangfeng~Cliff Zhang.
\newblock Online single-cell data integration through projecting heterogeneous datasets into a common cell-embedding space.
\newblock \emph{Nature Communications}, 13\penalty0 (1):\penalty0 6118, 2022.
\newblock \doi{10.1038/s41467-022-33758-z}.

\bibitem[Yang et~al.(2022)Yang, Wang, Wang, Fang, Tang, Huang, Lu, et~al.]{yang2022scbert}
Fan Yang, Wenchuan Wang, Fang Wang, Yuan Fang, Duyu Tang, Junzhou Huang, Hui Lu, et~al.
\newblock scbert as a large-scale pretrained deep language model for cell type annotation of single-cell {RNA}-seq data.
\newblock \emph{Nature Machine Intelligence}, 4:\penalty0 852--866, 2022.
\newblock \doi{10.1038/s42256-022-00534-z}.

\bibitem[Yang et~al.(2021)Yang, Belyaeva, Venkatachalapathy, Damodaran, Katcoff, Radhakrishnan, Shivashankar, and Uhler]{yang2021cmae}
Karren~Dai Yang, Anastasiya Belyaeva, Saradha Venkatachalapathy, Karthik Damodaran, Abigail Katcoff, Adityanarayanan Radhakrishnan, G.~V. Shivashankar, and Caroline Uhler.
\newblock Multi-domain translation between single-cell imaging and sequencing data using autoencoders.
\newblock \emph{Nature Communications}, 12\penalty0 (1):\penalty0 31, 2021.
\newblock \doi{10.1038/s41467-020-20249-2}.

\bibitem[Yuan and Kelley(2022)]{yuan2022scbasset}
Han Yuan and David~R. Kelley.
\newblock {scBasset}: Sequence-based modeling of single-cell {ATAC}-seq using convolutional neural networks.
\newblock \emph{Nature Methods}, 19\penalty0 (9):\penalty0 1088--1096, 2022.
\newblock \doi{10.1038/s41592-022-01562-8}.

\bibitem[Zhang et~al.(2024)Zhang, Zemke, Armand, and Ren]{zhang2024snapatac2}
Kai Zhang, Nathan~R. Zemke, Ethan~J. Armand, and Bing Ren.
\newblock A fast, scalable and versatile tool for analysis of single-cell omics data.
\newblock \emph{Nature Methods}, 21\penalty0 (2):\penalty0 217--227, 2024.
\newblock \doi{10.1038/s41592-023-02139-9}.

\bibitem[Zheng et~al.(2017)Zheng, Terry, Belgrader, Ryvkin, Bent, Wilson, Ziraldo, Wheeler, McDermott, Zhu, et~al.]{zheng2017massively}
Grace~XY Zheng, Jessica~M Terry, Phillip Belgrader, Paul Ryvkin, Zachary~W Bent, Ryan Wilson, Solongo~B Ziraldo, Tobias~D Wheeler, Geoff~P McDermott, Junjie Zhu, et~al.
\newblock Massively parallel digital transcriptional profiling of single cells.
\newblock \emph{Nature communications}, 8\penalty0 (1):\penalty0 14049, 2017.

\bibitem[Zheng et~al.(2024)Zheng, Long, Lu, Yang, Dai, Zhang, Nie, Ma, and Zhou]{zheng2024esm}
Kangjie Zheng, Siyu Long, Tianyu Lu, Junwei Yang, Xinyu Dai, Ming Zhang, Zaiqing Nie, Wei-Ying Ma, and Hao Zhou.
\newblock {ESM} all-atom: Multi-scale protein language model for unified molecular modeling.
\newblock In \emph{Proceedings of the 41st International Conference on Machine Learning}, volume 235 of \emph{Proceedings of Machine Learning Research}, pages 61432--61453. PMLR, 2024.
\newblock URL \url{https://proceedings.mlr.press/v235/zheng24h.html}.

\bibitem[Zhou et~al.(2024)Zhou, Ji, Li, Dutta, Davuluri, and Liu]{zhou2024dnabert2}
Zhihan Zhou, Yanrong Ji, Weijian Li, Pratik Dutta, Ramana~V Davuluri, and Han Liu.
\newblock {DNABERT}-2: Efficient foundation model and benchmark for multi-species genomes.
\newblock In \emph{The Twelfth International Conference on Learning Representations}, 2024.
\newblock URL \url{https://openreview.net/forum?id=oMLQB4EZE1}.

\end{thebibliography}
\clearpage
\appendix

\section{Related Works}
\label{sec::related_works}

\subsection{Foundation Models for Genomic Sequences}

Recent foundation models in genomics have increasingly focused on learning representations directly from genomic sequences, that is, modeling how DNA sequence encodes transcriptional and regulatory function. A first line of work is sequence-to-function modeling, in which models are trained to predict functional readouts from sequence windows. Borzoi is a representative example in this family: given a DNA sequence as input, it predicts RNA-seq coverage and provides a unified view of transcriptional and post-transcriptional regulation \cite{linder2025borzoi}. AlphaGenome extends this paradigm to a broader multimodal setting by taking long DNA sequence context as input and predicting a diverse collection of genomic outputs, including gene expression, transcription initiation, chromatin accessibility and so on \cite{avsec2026alphagenome}. GET further connects sequence-anchored regulatory features to transcription by combining chromatin accessibility and sequence information to predict gene expression across seen and unseen human cell types, while also supporting the inference of regulatory elements and regulators \cite{fu2025get}.

Scooby explicitly brings sequence-to-function modeling to single-cell resolution by equipping Borzoi with a cell-specific decoder~\citep{hingerl2025scooby}. This is a closely related approach; \method{} instead constructs shared peak and gene representations and integrates them through cell-specific multi-scale interactions. We include scooby in the expression benchmark. NT-v3 jointly models sequence and function and provides an alternative sequence encoder for our ablation~\citep{ntv3_2025}.

Another line of work is sequence language modeling, represented by Evo2, which is trained generatively on DNA sequences and aims to capture sequence semantics useful for tasks such as sequence analysis, variant scoring, and sequence design \cite{brixi2025evo2}. However, despite their strong performance, current sequence-based foundation models still primarily operate at local or regional sequence scales. For example, Borzoi uses 524-kb input sequences, whereas AlphaGenome and Evo2 operate on contexts of about 1~Mb \cite{linder2025borzoi,avsec2026alphagenome,brixi2025evo2}. Such contexts are already long by current standards, but they remain far from the scale required to directly capture more global regulatory semantics across large genomic neighborhoods or long-range transcriptional dependencies. This limits their ability to model broader regulatory programs beyond local sequence context alone.

\subsection{Foundation Models for Genes and Regulatory Regions}

A complementary family of models is built on single-cell omics data, where the primary modeling units are genes in scRNA-seq or accessible regions (for example, peaks) in scATAC-seq. In the scRNA setting, recent foundation models mainly learn gene-level semantics by treating each cell as a structured collection or sequence of gene tokens. Geneformer uses self-supervised pretraining to learn context-dependent gene-network representations from large-scale single-cell transcriptomes \cite{theodoris2023geneformer}. scGPT adopts a generative pretrained transformer framework for single-cell biology and provides a unified backbone for annotation, integration, perturbation prediction, and other downstream tasks \cite{cui2024scgpt}. scFoundation further scales this paradigm with a large pretrained model over more than 50 million human single-cell transcriptomes and emphasizes modeling long-range gene context within a cell \cite{hao2024scfoundation}. scBERT follows a BERT-style pretraining-and-finetuning scheme to capture gene--gene interactions from large-scale unlabeled scRNA-seq data, originally with a strong focus on robust cell type annotation \cite{yang2022scbert}.

Focussing on chromatin-based models, recent models have begun to learn regulatory representations from accessible-region-level signals. ChromFound is a dedicated foundation model for scATAC-seq that uses genome-aware tokenization and a hybrid architecture to learn universal chromatin representations from large-scale accessibility data, supporting zero-shot cell representation learning, transfer learning, and cross-omics prediction \cite{jiao2025chromfound}. EpiAgent similarly learns single-cell epigenomic representations through contextual modeling of accessibility profiles~\citep{chen2025epiagent}; we include it as a cross-omics prediction baseline. These fixed-vocabulary accessibility representations use peaks as primary units, while largely ignoring the fact that peaks and genes are ultimately grounded in the underlying DNA sequence. In particular, two different peaks may be treated as independent tokens once converted into accessibility features, even though they may share related sequence grammar, motif composition, or cis-regulatory logic. This missing sequence-level linkage is precisely the gap that motivates our model design.

\FloatBarrier
\section{Details of Pre-training}
\label{app:pretrain_details}

We pre-train \method{} on PBMC169K using the peak masked language modeling objective. AlphaGenome embeddings are used as fixed sequence features, and the input and output embeddings are tied for masked peak prediction. The full pre-training configuration is summarized in Table~\ref{tab:pretrain_hyper}.

\begin{table}[htbp]
\footnotesize
\centering
\caption{\method{} hyperparameters for peak MLM pre-training.}
\label{tab:pretrain_hyper}
\begin{tabular}{lc}
\toprule
\textbf{Hyperparameter} & \textbf{Value} \\
\midrule
Training objective & Peak MLM \\
Dataset & PBMC169K \\
Input sequence embedding & AlphaGenome embedding \\
Input dimension & 3072 \\
Model dimension & 2048 \\
Encoder layer types & 1 layer of peak self-attention, 2 layers of  cross interaction blocks \\
Number of attention heads & 16 \\
Maximum peaks per cell & 10,000 \\
Peak--gene distance bias & Enabled \\
Distance-based reweighting & Enabled \\
Distance mask value & -35 \\
Peak Mask Ratio & 15\% \\
Optimizer & Adam \\
Adam $(\beta_1,\beta_2)$ & $(0.9,0.98)$ \\
Adam $\epsilon$ & $10^{-6}$ \\
Learning rate & $10^{-3}$ \\
LR scheduler & inverse\_sqrt \\
Warmup updates & 100 \\
Gradient accumulation steps & 8 \\
Batch size (per GPU) & 1 \\
Training epochs & 5 \\
Hardware & 4$\times$ H200 GPUs \\
Training time & $\sim$45 hours \\
\bottomrule
\end{tabular}
\end{table}

\FloatBarrier
\section{Details of Gene Expression Prediction Task}
\label{app:gex_pred_details}

\subsection{Baseline Training Protocol}

For GET and EpiBERT, which are evaluated at the bulk or pseudobulk level, we first aggregate training cells from seen cell types into pseudobulk profiles. We then remove peaks and genes located in the held-out sequence folds and train the models only on the remaining seen-sequence regions. For the Borzoi and AlphaGenome baselines, which are evaluated without a cell-specific decoder, we construct cell-type pseudobulk profiles for each dataset and fine-tune a gene-level celltype-specific prediction head on top of the sequence encoder. The input to the head is the sequence representation of each gene, and the output is the corresponding gene expression count. In the unseen-cell-type setting, the prediction head is trained only on seen cell types and directly applied to held-out cell types during testing.

\subsection{Evaluation Gene Sets}

We select 5,257 highly variable genes from PBMC10K as the initial evaluation gene set, among which 1,161 genes are located in the held-out sequence folds and are unseen during training. For other datasets, we intersect this gene list with the genes available in each dataset. This results in 3,762 genes for BMMC, 4,948 genes for Openproblem, and all 5,257 genes for Bone.

\subsection{Held-out Cell Types}

For each dataset, we define an unseen-cell-type split by holding out selected cell types that are less similar to the training populations. These cell types are excluded from training and used only for testing.

\begin{table}[htbp]
\centering
\small
\caption{Held-out cell types used in the unseen-cell-type setting.}
\label{tab:heldout_celltypes}
\begin{tabular}{ll}
\toprule
\textbf{Dataset} & \textbf{Held-out cell types} \\
\midrule
PBMC10K &
TEM/TEMRA cytotoxic T cells; regulatory T cells; classical monocytes; \\
& non-classical monocytes; DC2; pDC \\
\midrule
BMMC &
Erythroblast; proerythroblast; normoblast; CD16$^+$ mono; \\
& CD8$^+$ T naive; pDC; ID2-hi myeloid progenitor; plasma cell; ILC \\
\midrule
Openproblem &
Erythrocyte progenitor; megakaryocyte progenitor; \\
& monocyte progenitor; B-cell progenitor \\
\midrule
Bone &
ChondroPro2; MandibularChon; SutureMes1; osteoblast; \\
& lymphatic endothelial; Schwann; erythroid; arachnoid; PerineuralFIB \\
\bottomrule
\end{tabular}
\end{table}

\subsection{\method{} Fine-tuning Configuration}

For gene expression prediction, we initialize \method{} from the peak MLM checkpoint and fine-tune it with the Poisson gene expression prediction objective. The main single-run benchmark uses seed 1; the additional BMMC robustness study uses seeds 0--4 (Appendix~\ref{app:multiseed_gex}). We use the same architecture and optimizer settings across datasets, with dataset-specific training epochs and batch sizes summarized in Table~\ref{tab:gex_finetune_hyper}.

\begin{table}[htbp]
\footnotesize
\centering
\caption{\method{} hyperparameters for gene expression prediction fine-tuning.}
\label{tab:gex_finetune_hyper}
\begin{tabular}{lc}
\toprule
\textbf{Hyperparameter} & \textbf{Value} \\
\midrule
Initialization & Peak MLM checkpoint \\
Input sequence embedding & AlphaGenome embedding \\
Input dimension & 3072 \\
Model dimension & 2048 \\
Encoder layer types & 1 layer of peak self-attention, 2 layers of  cross interaction blocks \\
Number of attention heads & 16 \\
Maximum peaks per cell & 10,000 \\
Peak--gene distance bias & Enabled \\
Distance-based reweighting & Enabled \\
Distance mask value & -35 \\
Loss function & Poisson loss \\
Optimizer & Adam \\
Adam $(\beta_1,\beta_2)$ & $(0.9,0.98)$ \\
Adam $\epsilon$ & $10^{-6}$ \\
Learning rate & $10^{-4}$ \\
LR scheduler & inverse\_sqrt \\
Warmup updates & 100 \\
Gradient accumulation steps & 4 \\
PBMC10K epochs / batch size & 30 / 8 \\
BMMC epochs / batch size & 30 / 8 \\
Openproblem epochs / batch size & 15 / 8 \\
HCA Bone epochs / batch size & 5 / 16 \\
\bottomrule
\end{tabular}
\end{table}

\subsection{Evaluation Protocol}

For the main comparison in Table~\ref{tab:cross_omics_prediction_main}, evaluation is performed at the cell-type pseudobulk level. The separate single-cell evaluation is reported in Appendix~\ref{app:single_cell_gex}. Pseudobulk aggregation is conducted separately within each split: for bulk or pseudobulk baselines, training targets are constructed by aggregating cells within the training split, while evaluation is performed on pseudobulk profiles aggregated only from the test split. This prevents information leakage across splits and enables a fair comparison between single-cell models and bulk or pseudobulk sequence models. Before metric computation, gene expression values are normalized as $\log_{10}(\mathrm{CPM}+1)$.

Let $y\in\mathbb{R}^{N}$ and $\hat y\in\mathbb{R}^{N}$ denote the observed and predicted pseudobulk expression profiles over $N$ evaluated genes for a given cell type. We report Pearson correlation coefficient (PCC), concordance correlation coefficient (CCC), and $R^2$. PCC measures linear correlation:
\[
\mathrm{PCC}(y,\hat y)
=
\frac{\sum_{i=1}^{N}(y_i-\bar y)(\hat y_i-\bar{\hat y})}
{\sqrt{\sum_{i=1}^{N}(y_i-\bar y)^2}\sqrt{\sum_{i=1}^{N}(\hat y_i-\bar{\hat y})^2}}.
\]
CCC further accounts for both correlation and agreement in scale and mean:
\[
\mathrm{CCC}(y,\hat y)
=
\frac{2\rho\sigma_y\sigma_{\hat y}}
{\sigma_y^2+\sigma_{\hat y}^2+(\bar y-\bar{\hat y})^2},
\]
where $\rho$ is the PCC between $y$ and $\hat y$, and $\sigma_y$ and $\sigma_{\hat y}$ are their standard deviations. $R^2$ measures the fraction of variance explained by the prediction:
\[
R^2(y,\hat y)
=
1-\frac{\sum_{i=1}^{N}(y_i-\hat y_i)^2}
{\sum_{i=1}^{N}(y_i-\bar y)^2}.
\]
Final results are averaged over cell types within each evaluation split.

\subsection{Baseline Models}
\label{app:baseline_models}

\paragraph{GET.}
GET~\cite{fu2025get} is a general expression transformer for modeling transcriptional regulation across human cell types. It takes chromatin accessibility and regulatory sequence-derived features as input, organized over a local genomic region around each gene, and learns transcriptional regulatory syntax through masked pretraining on pseudobulk scATAC-seq profiles. During fine-tuning, GET maps the learned chromatin regulatory representation to gene expression, enabling expression prediction in both seen and unseen cell types. In our experiments, GET is used as an ATAC+Seq baseline and initialized from the pretrained checkpoint released by the authors.

\paragraph{ChromFound.}
ChromFound~\cite{jiao2025chromfound} is a foundation model tailored for single-cell chromatin accessibility data. It takes single-cell scATAC-seq profiles as input, represents open chromatin regions using genome-aware tokenization, and uses a hybrid architecture to capture genome-wide regulatory context from sparse chromatin accessibility profiles. The model outputs cell embeddings that can be transferred to downstream tasks such as cell-type annotation and cross-omics prediction. In our experiments, ChromFound is used as a non-sequence single-cell baseline and initialized from the pretrained checkpoint released by the authors.

\paragraph{CMAE.}
CMAE~\cite{yang2021cmae} is a cross-modal autoencoder framework for translating between different single-cell data modalities. It uses modality-specific encoders and decoders to map each modality into a shared latent space, with latent alignment enabling translation between modalities that are not necessarily measured in the same cell. For scATAC-seq to scRNA-seq prediction, CMAE takes chromatin accessibility features as input and decodes the shared latent representation into gene expression profiles. In our experiments, CMAE is used as a non-sequence cross-modal baseline.

\paragraph{scMoGNN.}
scMoGNN~\cite{wen2022scmognn} is a graph neural network framework for multimodal single-cell data integration. It constructs a cell-feature bipartite graph from a given modality, where cells and biological features such as genes, proteins, or ATAC peaks are represented as graph nodes. The model performs cell-feature graph convolution to propagate information between cells and features, and then applies a task-specific head for modality prediction, matching, or joint embedding. In our experiments, scMoGNN takes single-cell ATAC features as input and predicts gene expression profiles.

\paragraph{AlphaGenome.}
AlphaGenome~\cite{avsec2026alphagenome} is a unified DNA sequence-to-function model for regulatory genomics. It takes long DNA sequences as input and predicts thousands of functional genomic tracks across modalities including gene expression, transcription initiation, chromatin accessibility, histone modifications, transcription factor binding, chromatin contacts, and splicing-related signals. For gene expression prediction, we use sequence-derived representations from AlphaGenome and fine-tune a gene-level prediction head. In our experiments, AlphaGenome is implemented using the PyTorch reimplementation at \url{https://github.com/genomicsxai/alphagenome-pytorch} and initialized from released pretrained weights.

\paragraph{EpiBERT.}
EpiBERT~\cite{javed2025epibert} is a multi-modal transformer for cell-type-agnostic regulatory prediction. It integrates genomic sequence with cell-type-specific chromatin accessibility and is pretrained using a masked accessibility modeling objective inspired by BERT. The learned representations can be fine-tuned for downstream regulatory tasks, including gene expression prediction in held-out genes and cell types. In our experiments, EpiBERT is used as an ATAC+Seq baseline and initialized from the pretrained checkpoint released by the authors.

\paragraph{Borzoi.}
Borzoi~\cite{linder2025borzoi} is a sequence-based model for predicting RNA-seq coverage and other regulatory genomic assays directly from DNA sequence. It models long genomic contexts to predict base- or bin-level coverage tracks, including RNA-seq, CAGE, DNase-seq, ATAC-seq, and ChIP-seq signals, thereby capturing multiple layers of gene regulation such as transcription, splicing, and polyadenylation. For gene expression prediction, we aggregate Borzoi sequence-derived outputs or representations to the gene level. In our experiments, Borzoi is used as a sequence-only baseline and initialized from the pretrained checkpoint released by the authors.

\paragraph{Scooby, EpiAgent, and PCA+Ridge.}
Scooby~\citep{hingerl2025scooby} extends sequence-based prediction to single-cell resolution through a cell-specific decoder. Its pseudobulk benchmark results are reported in Appendix~\ref{app:additional_gex}. EpiAgent~\citep{chen2025epiagent} learns representations from single-cell epigenomic profiles; for cross-omics prediction we fine-tune it following the ChromFound protocol. PCA+Ridge provides a linear scATAC-to-scRNA baseline on PBMC10K and BMMC (Appendix~\ref{app:simple_gex}). These additions do not change the shared evaluation gene sets or the main benchmark's held-out sequence and cell-type definitions.

\FloatBarrier
\section{Details of Cell Clustering Evaluation}
\label{app:cell_clustering_details}

We perform zero-shot cell clustering on the held-out PBMC169K batch \texttt{BIO\_ddseq\_1}. We follow the evaluation protocol of ChromFound~\citep{jiao2025chromfound}, and some baseline results in Table~\ref{tab:cell_clustering_pbmc169k} are taken from their paper. For each method, we compute cell embeddings from scATAC-seq data, followed by PCA with 50 components and K-means clustering with 10 initializations. We repeat the full clustering pipeline with 10 random seeds and report the mean and standard deviation of ARI, FMI, NMI, and AMI.

For single-cell models, we directly use their learned cell embeddings. For bulk sequence models, which do not produce cell-level representations, we first compute peak embeddings and then average the embeddings of all accessible peaks within each cell to obtain a cell embedding.

\subsection{Evaluation Metrics}

Let $Y$ denote the ground-truth cell-type labels and $C$ denote the predicted cluster labels. For pair-counting metrics, we define TP, TN, FP, and FN as the numbers of cell pairs that are correctly grouped together, correctly separated, incorrectly grouped together, and incorrectly separated, respectively.

The adjusted Rand index (ARI) corrects the Rand index for chance:
\[
\mathrm{RI}
=
\frac{\mathrm{TP}+\mathrm{TN}}
{\mathrm{TP}+\mathrm{TN}+\mathrm{FP}+\mathrm{FN}},
\qquad
\mathrm{ARI}
=
\frac{\mathrm{RI}-\mathbb{E}[\mathrm{RI}]}
{\max(\mathrm{RI})-\mathbb{E}[\mathrm{RI}]}.
\]
Higher ARI indicates better agreement between predicted clusters and cell-type labels after accounting for random clustering.

The Fowlkes--Mallows index (FMI) measures the geometric mean of pairwise precision and recall:
\[
\mathrm{FMI}
=
\sqrt{
\frac{\mathrm{TP}}{\mathrm{TP}+\mathrm{FP}}
\cdot
\frac{\mathrm{TP}}{\mathrm{TP}+\mathrm{FN}}
}.
\]
A higher FMI indicates that cells assigned to the same cluster are more consistently from the same ground-truth cell type.

For information-theoretic metrics, we use mutual information between $Y$ and $C$. The normalized mutual information (NMI) is computed as
\[
\mathrm{NMI}(Y,C)
=
\frac{2I(Y;C)}{H(Y)+H(C)},
\]
where $I(Y;C)$ is the mutual information and $H(\cdot)$ is entropy. NMI measures the shared information between predicted clusters and cell-type labels after normalization.

The adjusted mutual information (AMI) further corrects mutual information for chance:
\[
\mathrm{AMI}(Y,C)
=
\frac{I(Y;C)-\mathbb{E}[I(Y;C)]}
{\frac{1}{2}\big(H(Y)+H(C)\big)-\mathbb{E}[I(Y;C)]}.
\]
Higher AMI indicates stronger label--cluster agreement after accounting for the mutual information expected under random assignments.

\FloatBarrier
\subsection{PCA baseline for zero-shot cell clustering}
\label{app:pca_clustering}
We apply PCA directly to the same binary scATAC input from the held-out PBMC169K batch \texttt{BIO\_ddseq\_1} and cluster the resulting cell representations. Table~\ref{tab:pca_clustering} compares this baseline with AlphaGenome, ChromFound, and \method{}. PCA improves on the underlying sequence-encoder baseline, but \method{} retains higher scores on all four metrics.

\begin{table}[htbp]
\centering
\caption{PCA comparison for zero-shot cell clustering. Values are mean $\pm$ standard deviation where reported; PCA standard deviations are available only for ARI.}
\label{tab:pca_clustering}
\footnotesize
\setlength{\tabcolsep}{5pt}
\renewcommand{\arraystretch}{1.12}
\begin{tabular}{lcccc}
\toprule
Model & ARI & FMI & NMI & AMI \\
\midrule
AlphaGenome & $0.1032\pm0.0000$ & $0.3814\pm0.0000$ & $0.1646\pm0.0000$ & $0.1633\pm0.0000$ \\
ChromFound & $0.4835\pm0.0082$ & $0.6604\pm0.0042$ & $0.5950\pm0.0036$ & $0.5944\pm0.0036$ \\
PCA & $0.1649\pm0.0221$ & 0.4473 & 0.2856 & 0.2845 \\
\rowcolor{mygray}
\method{} & $\mathbf{0.5377\pm0.0008}$ & $\mathbf{0.6973\pm0.0006}$ & $\mathbf{0.6085\pm0.0003}$ & $\mathbf{0.6079\pm0.0003}$ \\
\bottomrule
\end{tabular}
\end{table}
\FloatBarrier

\section{PBMC10K Few-Shot Evaluation Protocol}
\label{app:pbmc10k_fewshot}

To evaluate the data efficiency of \method{}, we performed a few-shot study on the PBMC10K gene expression prediction task. All few-shot experiments were initialized from the same MLM-pretrained checkpoint. For each training-set size, we fine-tuned the model on a stratified subset of the PBMC10K training split and evaluated the resulting checkpoint using the same PBMC10K unseen-sequence evaluation pipeline as in the main experiments.

The training subsets contained
\[
N \in \{0, 1, 2, 4, 8, 16, 32, 64, 128, 256, 512, 1024, 2048, 4096, \mathrm{all}\}
\]
paired training cells, where the all-data setting contains 5,649 cells. For each nonzero few-shot setting, cells were sampled from the training split using stratified-by-cell-type subsampling with a fixed random seed. Validation and test splits were not subsampled; all evaluations were performed on the full PBMC10K validation-test split. The evaluation cell types were also kept fixed across all settings, including \texttt{TEM\_TEMRA\_CYTOTOXIC\_T\_CELLS}, \texttt{REGULATORY\_T\_CELLS}, \texttt{CLASSICAL\_MONOCYTES}, \texttt{NON-CLASSICAL\_MONOCYTES}, \texttt{DC2}, and \texttt{PDC}.

All runs used the same optimization configuration unless otherwise specified. We used Adam with $\beta=(0.9,0.98)$ and $\epsilon=10^{-6}$, an inverse-square-root learning-rate scheduler with 50 warmup updates, learning rate $10^{-4}$, gradient clipping norm 0.0. The batch size was set to 1. Therefore, the number of optimizer steps per epoch is equal to the number of training cells in the corresponding few-shot setting. We used \texttt{max-cell-peaks-num} of 10,000 and the poisson loss as gene expression prediction loss. For unseen-sequence evaluation, genes in folds \texttt{fold3} and \texttt{fold4} were held out \cite{linder2025borzoi}, which is the same with all other unseen sequence settings.

The epoch budget was chosen so that the total number of optimizer steps increases monotonically with the number of training cells, while keeping the maximum epoch count no larger than 30 once $N \geq 256$. In this way, larger few-shot subsets receive more total optimization steps, which helps ensure sufficient convergence as the amount of supervision increases, rather than prematurely stopping models trained with more data. The exact training schedule is summarized in Table~\ref{tab:pbmc10k_fewshot_schedule}.

\begin{table}[htbp]
\centering
\caption{Training schedule for the PBMC10K few-shot evaluation. Since the batch size is 1, the total number of optimizer steps is equal to $N \times$ epochs.}
\label{tab:pbmc10k_fewshot_schedule}
\begin{tabular}{r r r}
\toprule
Training cells $N$ & Epochs & Total optimizer steps \\
\midrule
1 & 200 & 200 \\
2 & 200 & 400 \\
4 & 200 & 800 \\
8 & 150 & 1,200 \\
16 & 100 & 1,600 \\
32 & 80 & 2,560 \\
64 & 60 & 3,840 \\
128 & 50 & 6,400 \\
256 & 30 & 7,680 \\
512 & 30 & 15,360 \\
1,024 & 30 & 30,720 \\
2,048 & 30 & 61,440 \\
4,096 & 30 & 122,880 \\
All (5,649) & 30 & 169,470 \\
\bottomrule
\end{tabular}
\end{table}

For checkpointing, each run was configured to retain only the final checkpoint. Specifically, we kept the last epoch only, disabled the default last-checkpoint writing, used a large save interval, and disabled validation during training to avoid automatic best-checkpoint generation. Any remaining \texttt{checkpoint\_best.pt} file was removed after training. Thus, each few-shot run produced exactly one checkpoint, corresponding to the final training epoch.

For the zero-shot setting ($N=0$), no task-specific fine-tuning was performed. Instead, we directly attached a randomly initialized gene-expression prediction head to the peak-MLM-pretrained model and evaluated it using the same PBMC10K pipeline. Since this prediction head is not trained on any paired expression data, the resulting predictions are expected to be largely uninformative, which explains the near-zero PCC observed in this setting.

\FloatBarrier
\section{Details of biological application methodology }
\label{app:bio_app}

We construct a motif-informed benchmark to evaluate whether models can prioritise regulatory interactions linking transcription factors (TFs), open chromatin regions (peaks), and target genes using DNA sequence.

\subsection{Dataset and candidate interactions}

We use open chromatin peaks from PBMC data and curated TF–gene regulatory interactions from CollecTRI~\cite{muller2023collectri}. This contains 43,175 signed TF–gene interactions across 1,186 TFs. Candidate regulatory interactions are defined by assigning peaks to genes within a 2 Mb cis window around the transcription start site, excluding promoter-proximal regions (<2 kb). Peaks are grouped into distance bins (2–50 kb, 50–100 kb, 100–250 kb, 250 kb–2 Mb).

Peak sequences are scanned for TF motifs using FIMO from the MEME Suite \citep{grant2011fimo, bailey2015meme} with CollecTRI-overlapping JASPAR position weight matrices \citep{fornes2020jaspar, muller2023collectri}, calling instances at $p \le 1\times10^{-5}$ and $q \le 0.1$ (FDR-controlled). This yields 2.3k--44.7k positive gene--peak pairs across 411--668 genes (increasing with distance), with 132k--794k peak--TF motif hits per bin after filtering. The retained motif set spans $\sim$135--165 TFs per bin, with no single factor dominating (top $\sim$5--9\% of hits; top five $\sim$25--30\%).

\subsection{Positive and control definitions}

For each gene--peak pair, we consider the TFs annotated by CollecTRI as regulators of that gene and identify which have motif support within the peak.

The broad setting requires a motif for at least one relevant TF. The conservative setting requires motifs for at least two distinct relevant TFs and excludes positives supported only by GC-rich or ubiquitous TF families (e.g., SP/KLF). Motif instances are not removed from peaks; rather, positive labels supported exclusively by these families are excluded. This provides a stricter test of regulatory specificity beyond motif frequency and GC content biases.

Controls are constructed within the same gene and distance bin to control for gene identity and genomic distance. A control peak–gene pair is required to (i) contain no motifs for any CollecTRI TF annotated to regulate that gene, and (ii) contain at least one motif for other transcription factors. This ensures that comparisons are not driven by the generic presence or absence of motifs, but instead test whether models prioritise peaks containing motifs for the correct regulators of each gene.

This formulation explicitly accounts for the fact that genes may have multiple known regulators: positives may be supported by any of these TFs, while controls must exclude all of them.

For each positive pair, we sample up to five matched controls from peaks assigned to the same gene and distance bin. This defines matched groups consisting of one positive and its associated controls. Groups lacking sufficient controls are excluded. This matching strategy ensures that comparisons are performed within comparable genomic contexts.

\subsection{Evaluation metrics}

We evaluate model performance using two complementary metrics.

\textbf{Top-\textit{k} enrichment.}

 For each gene and distance bin, peaks are ranked by model score (cosine similarity between peak and gene embeddings, or baseline scores). We deduplicate to unique gene–peak–distance-bin pairs before ranking. Enrichment is computed per gene as the number of positives observed among the top-\textit{k} ranked peaks divided by the expected number under the gene-specific positive rate, and results are reported as the mean across genes.

\textbf{Matched pairwise AUROC (win-rate).}

 Using matched groups, we compare each positive score against those of its matched controls (ties scored as 0.5). Group-level win-rates are averaged per gene, and final values are reported as the mean across genes. This formulation ensures that each gene contributes equally and avoids biases from varying numbers of candidate peaks.

For both metrics, genes lacking at least one positive and one control, or insufficient candidates for evaluation, are excluded. Error bars denote 95\% confidence intervals across genes, computed using the standard error of the mean.

\subsection{Additional considerations}

Baseline sequence models operate on fixed local input windows (typically $\sim$1\,Mb) and encode positional information only within each window, without an explicit representation of global genomic coordinates. In line with their intended usage, we tile the genome into non-overlapping windows for scalable inference; however, this implies that nearby peaks and genes may be processed in different windows and therefore lack shared contextual representation, potentially reducing embedding similarity despite genomic proximity. In contrast, our model incorporates genome-wide positional context, enabling more consistent representation across windows, which may contribute to improved long-range regulatory prioritisation. To account for this and to be fair to other models, we focus evaluation of our performance relative to distance-matched and random baselines, ensuring that improvements reflect signal beyond proximity alone.

We note additional limitations of the benchmark: CollecTRI{\cite{muller2023collectri}} integrates heterogeneous sources of TF--gene regulation and may include indirect or context-specific interactions, and motif presence does not imply TF binding or activity. Incorporating orthogonal evidence such as chromatin accessibility footprints {\cite{baek2017bivariate}} may further refine these labels.

\clearpage
\section{Additional Gene Expression Evaluations}
\label{app:additional_gex}

\subsection{Additional pseudobulk baseline: EpiAgent}
Table~\ref{tab:gex_epiagent} compares EpiAgent~\citep{chen2025epiagent} and \method{} on seen sequences. Normalization and splits follow Appendix~\ref{app:gex_pred_details}; EpiAgent uses the ChromFound fine-tuning protocol. \method{} achieves higher PCC, CCC, and $R^2$ across all four datasets and both cell-type settings. Unseen-sequence diagnostics are reported in Section~\ref{app:simple_gex}.

\begin{table}[htbp]
\centering
\caption{EpiAgent comparison on seen sequences. Entries are PCC / CCC / $R^2$; CT denotes the cell-type split.}
\label{tab:gex_epiagent}
\small
\setlength{\tabcolsep}{6pt}
\renewcommand{\arraystretch}{1.12}
\begin{tabular}{llcc}
\toprule
Dataset & CT & EpiAgent & \method{} \\
\midrule
PBMC10K & Seen & 0.763 / 0.524 / 0.092 & 0.909 / 0.895 / 0.811 \\
 & Unseen & 0.746 / 0.503 / $-0.009$ & 0.893 / 0.887 / 0.788 \\
\midrule
BMMC & Seen & 0.819 / 0.575 / 0.131 & 0.983 / 0.981 / 0.962 \\
 & Unseen & 0.818 / 0.620 / 0.248 & 0.948 / 0.937 / 0.873 \\
\midrule
Openproblem & Seen & 0.876 / 0.683 / 0.116 & 0.996 / 0.995 / 0.990 \\
 & Unseen & 0.860 / 0.669 / 0.079 & 0.979 / 0.978 / 0.956 \\
\midrule
Bone & Seen & 0.929 / 0.666 / $-0.137$ & 0.958 / 0.945 / 0.897 \\
 & Unseen & 0.915 / 0.680 / $-0.070$ & 0.940 / 0.928 / 0.863 \\
\bottomrule
\end{tabular}
\end{table}

\FloatBarrier

\subsection{Single-cell expression prediction}
\label{app:single_cell_gex}
Table~\ref{tab:gex_singlecell} reports PCC at single-cell resolution rather than after cell-type pseudobulk aggregation. \method{} is competitive on seen sequences: it leads on BMMC and Openproblem, while scMoGNN leads on PBMC10K and CMAE on Bone. On unseen sequences, \method{} retains positive predictive correlation on all four datasets, whereas the fixed-vocabulary comparison models have near-zero reported scores.

\begin{table}[htbp]
\centering
\caption{Single-cell gene expression PCC. Bold denotes the highest value within each sequence setting.}
\label{tab:gex_singlecell}
\small
\setlength{\tabcolsep}{10pt}
\begin{tabular}{lrrrr}
\toprule
Model & PBMC10K & BMMC & Openproblem & Bone \\
\midrule
\multicolumn{5}{c}{\textbf{Seen sequences}} \\
CMAE & 0.861 & 0.680 & 0.627 & \textbf{0.827} \\
scMoGNN & \textbf{0.903} & 0.720 & 0.689 & 0.805 \\
ChromFound & 0.837 & 0.658 & 0.534 & 0.774 \\
\method{} & 0.876 & \textbf{0.732} & \textbf{0.691} & 0.811 \\
\midrule
\multicolumn{5}{c}{\textbf{Unseen sequences}} \\
CMAE & $-0.014$ & $-0.007$ & 0.003 & $-0.000$ \\
scMoGNN & $-0.018$ & $-0.005$ & $-0.039$ & 0.014 \\
ChromFound & 0.010 & $-0.003$ & $-0.011$ & $-0.006$ \\
\method{} & \textbf{0.436} & \textbf{0.382} & \textbf{0.535} & \textbf{0.410} \\
\bottomrule
\end{tabular}
\end{table}
\FloatBarrier

\subsection{Five-seed robustness on BMMC}
\label{app:multiseed_gex}
We repeat the BMMC benchmark with seeds 0--4 and report PCC means and standard deviations in Table~\ref{tab:gex_multiseed}. \method{} remains the highest-scoring model in each setting, with standard deviations of 0.001--0.002. These repetitions supplement the seed-1 results in the main benchmark; fixed splits and deterministic pseudobulk aggregation alone do not quantify variation between training runs. The five-seed comparison is restricted to BMMC and the methods listed below.

\begin{table}[htbp]
\centering
\caption{BMMC pseudobulk PCC (mean $\pm$ standard deviation, five seeds). CT denotes cell types; dashes indicate unsupported unseen-sequence evaluation.}
\label{tab:gex_multiseed}
\small
\setlength{\tabcolsep}{6pt}
\renewcommand{\arraystretch}{1.15}
\begin{tabular}{lcccc}
\toprule
& \multicolumn{2}{c}{\textbf{Seen sequences}} & \multicolumn{2}{c}{\textbf{Unseen sequences}} \\
\cmidrule(lr){2-3}\cmidrule(lr){4-5}
Model & Seen CT & Unseen CT & Seen CT & Unseen CT \\
\midrule
CMAE & $0.463\pm0.038$ & $0.458\pm0.036$ & --- & --- \\
scMoGNN & $0.859\pm0.059$ & $0.827\pm0.049$ & --- & --- \\
ChromFound & $0.856\pm0.007$ & $0.822\pm0.007$ & --- & --- \\
Borzoi & $0.719\pm0.019$ & $0.690\pm0.016$ & $0.567\pm0.022$ & $0.550\pm0.016$ \\
AlphaGenome & $0.696\pm0.046$ & $0.650\pm0.058$ & $0.656\pm0.012$ & $0.628\pm0.020$ \\
GET & $0.597\pm0.008$ & $0.589\pm0.006$ & $0.205\pm0.017$ & $0.192\pm0.013$ \\
EpiBERT & $0.879\pm0.006$ & $0.853\pm0.006$ & $0.547\pm0.008$ & $0.561\pm0.008$ \\
\rowcolor{mygray}
\method{} & $\mathbf{0.982\pm0.001}$ & $\mathbf{0.946\pm0.001}$ & $\mathbf{0.747\pm0.002}$ & $\mathbf{0.740\pm0.001}$ \\
\bottomrule
\end{tabular}
\end{table}
\FloatBarrier

\subsection{PCA+Ridge and fixed-vocabulary failure diagnostics}
\label{app:simple_gex}
We include a PCA+Ridge linear model mapping scATAC inputs to scRNA outputs. It is a strong comparator for genes present during training (Table~\ref{tab:gex_linear}), reaching pseudobulk PCC 0.900 on PBMC10K and 0.972 on BMMC. Its output variables are nevertheless tied to the training gene set.

\begin{table}[htbp]
\centering
\caption{PCA+Ridge comparison. Entries are PCC / CCC / $R^2$.}
\label{tab:gex_linear}
\small
\setlength{\tabcolsep}{6pt}
\begin{tabular}{llcc}
\toprule
Dataset & Sequences & PCA+Ridge & \method{} \\
\midrule
PBMC10K & Seen & 0.900 / 0.878 / 0.782 & 0.909 / 0.895 / 0.811 \\
PBMC10K & Unseen & 0.000 / 0.000 / $-1.310$ & 0.684 / 0.663 / 0.409 \\
BMMC & Seen & 0.972 / 0.963 / 0.931 & 0.983 / 0.981 / 0.962 \\
BMMC & Unseen & 0.000 / 0.000 / $-1.707$ & 0.748 / 0.738 / 0.524 \\
\bottomrule
\end{tabular}
\end{table}

Table~\ref{tab:epiagent_unseen} similarly preserves the EpiAgent unseen-sequence diagnostics separately from the sequence-generalization benchmark. The near-zero correlations and negative $R^2$ values do not support generalization beyond its fixed vocabulary.
\begin{table}[htbp]
\centering
\caption{EpiAgent unseen-sequence failure diagnostics. CT denotes the cell-type split. These values do not represent native unseen-gene support.}
\label{tab:epiagent_unseen}
\small
\setlength{\tabcolsep}{10pt}
\begin{tabular}{llrrr}
\toprule
Dataset & CT & PCC & CCC & $R^2$ \\
\midrule
PBMC10K & Seen & -0.024 & -0.023 & -1.078 \\
 & Unseen & -0.013 & -0.012 & -1.115 \\
BMMC & Seen & -0.029 & -0.022 & -0.789 \\
 & Unseen & -0.020 & -0.015 & -0.861 \\
Openproblem & Seen & -0.026 & -0.024 & -0.665 \\
 & Unseen & -0.017 & -0.016 & -0.664 \\
Bone & Seen & -0.007 & -0.007 & -1.519 \\
 & Unseen & -0.011 & -0.011 & -1.553 \\
\bottomrule
\end{tabular}
\end{table}
\FloatBarrier

\clearpage
\section{Additional Architecture and Pretraining Ablations}
\label{app:additional_ablations}

\subsection{Two-sided versus upstream-only genomic context}
\label{app:two_sided_context}

Restricting each gene to upstream peaks reduces seen PCC and CCC, while improving unseen PCC at the cost of a slight decrease in unseen CCC (Table~\ref{tab:upstream_ablation}). These results highlight a trade-off in spatial context. Overall, bidirectional cross-scale information exchange remains the more effective design, while restricting the spatial context to upstream peaks does not yield consistent improvements.

\begin{table}[htbp]
\centering
\small
\setlength{\tabcolsep}{4pt}
\renewcommand{\arraystretch}{1.12}
\caption{PBMC10K comparison of two-sided and upstream-only genomic context using AlphaGenome. Bold marks the better value per metric.}
\label{tab:upstream_ablation}
\begin{tabular}{l|cc|cc}
\toprule
\textbf{Genomic context}
& \multicolumn{2}{c|}{\cellcolor{unseenpeach}\textbf{Unseen seq.}}
& \multicolumn{2}{c}{\cellcolor{seenblue}\textbf{Seen seq.}} \\
\cmidrule(lr){2-3}\cmidrule(lr){4-5}
& PCC$\uparrow$ & CCC$\uparrow$ & PCC$\uparrow$ & CCC$\uparrow$ \\
\midrule
\rowcolor{mygray}
Both sides (\method{}) & 0.684 & \textbf{0.663} & \textbf{0.909} & \textbf{0.895} \\
Upstream only & \textbf{0.701} & 0.659 & 0.885 & 0.864 \\
\bottomrule
\end{tabular}
\end{table}
\FloatBarrier

\subsection{Peak-MLM ablations across downstream tasks}
\label{app:peak_mlm_transfer}
\paragraph{Zero-shot cell clustering.}
Removing peak MLM reduces held-out clustering ARI from 0.5377 to 0.1221, approaching the AlphaGenome baseline (Table~\ref{tab:mlm_clustering}). The cross-task improvement supports learning cell-level accessibility context beyond the frozen sequence features.

\begin{table}[htbp]
\centering
\caption{Peak-MLM ablation for zero-shot cell clustering. Entries are means $\pm$ standard deviations.}
\label{tab:mlm_clustering}
\footnotesize
\setlength{\tabcolsep}{5pt}
\begin{tabular}{lcccc}
\toprule
Model & ARI & FMI & NMI & AMI \\
\midrule
AlphaGenome & $0.1032\pm0.0000$ & $0.3814\pm0.0000$ & $0.1646\pm0.0000$ & $0.1633\pm0.0000$ \\
\method{} w/o peak MLM & $0.1221\pm0.0006$ & $0.3773\pm0.0005$ & $0.2068\pm0.0007$ & $0.2056\pm0.0007$ \\
\method{} & $\mathbf{0.5377\pm0.0008}$ & $\mathbf{0.6973\pm0.0006}$ & $\mathbf{0.6085\pm0.0003}$ & $\mathbf{0.6079\pm0.0003}$ \\
\bottomrule
\end{tabular}
\end{table}

\paragraph{Enhancer--gene linking.}
Removing peak MLM lowers balanced AUROC, balanced AUPRC, and all reported eQTL retrieval metrics (Table~\ref{tab:mlm_linking}).

\begin{table}[htbp]
\centering
\caption{Peak-MLM ablation for eQTL enhancer--gene linking. ``Bal.'' denotes the balanced positive/negative evaluation. The last row is full minus ablated.}
\label{tab:mlm_linking}
\footnotesize
\setlength{\tabcolsep}{4pt}
\renewcommand{\arraystretch}{1.15}
\begin{tabular}{lrrrrrr}
\toprule
Model & \shortstack{AUROC\\(bal.)} & \shortstack{AUPRC\\(bal.)} & \shortstack{AUPRC\\micro} & \shortstack{AUPRC\\macro} & NDCG@10 & HR@10 \\
\midrule
\method{} & \textbf{0.6590} & \textbf{0.6499} & \textbf{0.1120} & \textbf{0.2051} & \textbf{0.2095} & \textbf{0.5571} \\
\method{} w/o peak MLM & 0.6106 & 0.6210 & 0.0422 & 0.1389 & 0.1479 & 0.3947 \\
Difference & 0.0484 & 0.0289 & 0.0698 & 0.0662 & 0.0616 & 0.1624 \\
\bottomrule
\end{tabular}
\end{table}
\FloatBarrier

\clearpage
\section{Enhancer--Gene Retrieval under Class Imbalance}
\label{app:retrieval}

\subsection{Per-gene retrieval with a broad candidate set}
The distance-matched AUROC experiment in Section~\ref{sec:peak_gene_enhancer_task} uses a balanced 1:1 positive-to-negative ratio. To evaluate a more imbalanced retrieval setting, we also rank candidate peaks for each gene within a $\pm1$ Mb cis window. eQTL-supported pairs define positives, and potential positives are excluded from the negative pool. The resulting positive-to-negative ratio is approximately 1:36.

We report micro- and macro-AUPRC together with normalized discounted cumulative gain (NDCG) and hit rate (HR) at ranks 5, 10, and 20. AUPRC evaluates precision--recall performance, while the top-ranked retrieval metrics assess the prioritization of experimentally supported candidates. As shown in Table~\ref{tab:retrieval_all}, \method{} has the highest score across all eight reported metrics among the compared sequence- and epigenome-based models.

\begin{table}[htbp]
\centering
\caption{eQTL enhancer--gene retrieval within $\pm1$ Mb, at approximately 1:36 class imbalance. Higher is better.}
\label{tab:retrieval_all}
\footnotesize
\setlength{\tabcolsep}{4pt}
\renewcommand{\arraystretch}{1.15}
\begin{tabular}{lrrrrrrrr}
\toprule
& \multicolumn{2}{c}{AUPRC} & \multicolumn{3}{c}{NDCG} & \multicolumn{3}{c}{HR} \\
\cmidrule(lr){2-3}\cmidrule(lr){4-6}\cmidrule(lr){7-9}
Model & Micro & Macro & @5 & @10 & @20 & @5 & @10 & @20 \\
\midrule
AlphaGenome & 0.0718 & 0.1363 & 0.1233 & 0.1403 & 0.1694 & 0.2999 & 0.4114 & 0.5421 \\
Borzoi & 0.0834 & 0.1581 & 0.1452 & 0.1673 & 0.1998 & 0.3488 & 0.4702 & 0.6052 \\
GET & 0.0357 & 0.1014 & 0.0971 & 0.1072 & 0.1288 & 0.2813 & 0.3982 & 0.5377 \\
EpiBERT & 0.0445 & 0.1016 & 0.0996 & 0.1081 & 0.1317 & 0.2847 & 0.3958 & 0.5245 \\
\rowcolor{mygray}
\method{} & \textbf{0.1271} & \textbf{0.2500} & \textbf{0.2331} & \textbf{0.2664} & \textbf{0.3157} & \textbf{0.4868} & \textbf{0.6492} & \textbf{0.7784} \\
\bottomrule
\end{tabular}
\end{table}

\subsection{A linear baseline and distance-stratified retrieval}
We fit a PCA+Ridge model from scATAC to scRNA and use its peak-by-gene coefficient matrix as the association score. This tests whether a linear cross-modal predictor is sufficient for the regulatory-linking task. Table~\ref{tab:retrieval_linear} compares the overall candidate set and three enhancer--gene distance bins.

\method{} improves overall micro-AUPRC from 0.0532 to 0.1271 and macro-AUPRC from 0.0920 to 0.2500. Its advantage is particularly pronounced beyond 100 kb, where micro-AUPRC rises from 0.0203 to 0.0528 and HR@10 from 0.3717 to 0.5932. The advantage is not universal at short distance: PCA+Ridge has higher HR@10 in the 0--50 kb bin (0.8677 versus 0.8478), while \method{} leads the other three metrics in that bin.

\begin{table}[htbp]
\centering
\caption{PCA+Ridge versus \method{} for eQTL retrieval. Bold indicates the higher value within each distance bin.}
\label{tab:retrieval_linear}
\small
\setlength{\tabcolsep}{6pt}
\renewcommand{\arraystretch}{1.15}
\begin{tabular}{llrrrr}
\toprule
Distance & Model & \shortstack{AUPRC\\micro} & \shortstack{AUPRC\\macro} & NDCG@10 & HR@10 \\
\midrule
All & PCA+Ridge & 0.0532 & 0.0920 & 0.1054 & 0.4056 \\
& \method{} & \textbf{0.1271} & \textbf{0.2500} & \textbf{0.2664} & \textbf{0.6492} \\
\midrule
0--50 kb & PCA+Ridge & 0.2008 & 0.3438 & 0.3955 & \textbf{0.8677} \\
& \method{} & \textbf{0.2039} & \textbf{0.3672} & \textbf{0.3974} & 0.8478 \\
\midrule
50--100 kb & PCA+Ridge & 0.1500 & 0.3872 & 0.4640 & 0.8690 \\
& \method{} & \textbf{0.1875} & \textbf{0.4859} & \textbf{0.5535} & \textbf{0.8946} \\
\midrule
$>100$ kb & PCA+Ridge & 0.0203 & 0.0985 & 0.1087 & 0.3717 \\
& \method{} & \textbf{0.0528} & \textbf{0.2537} & \textbf{0.2869} & \textbf{0.5932} \\
\bottomrule
\end{tabular}
\end{table}
\FloatBarrier

\clearpage
\section{Computational Efficiency and Peak-count Scaling}
\label{app:efficiency}

\subsection{Inference-time comparison}
We profile inference on the same 2,000 BMMC test cells using one NVIDIA H200 GPU (Table~\ref{tab:inference_time}). \method{} takes 61.18 ms per cell on average, compared with 5.94 ms for ChromFound and 0.99 ms for CMAE. The multi-scale representation therefore carries a substantial inference cost relative to lightweight baselines.

\begin{table}[htbp]
\centering
\caption{Inference on 2,000 BMMC test cells using one H200. Timing and throughput are reported with the precision of the measurements.}
\label{tab:inference_time}
\small
\setlength{\tabcolsep}{10pt}
\begin{tabular}{lrrr}
\toprule
Model & Mean (ms/cell) & Median (ms/cell) & Throughput (cells/s) \\
\midrule
CMAE & 0.99 & 0.99 & 1,006 \\
ChromFound & 5.94 & 5.93 & 168 \\
\method{} & 61.18 & 55.07 & 16.3 \\
\bottomrule
\end{tabular}
\end{table}

\subsection{Training memory with increasing peaks}
We profile simulated cells with more peaks using BF16 and a per-GPU batch size of one, with memory-efficient exact attention implemented through FlashAttention/FlexAttention. Table~\ref{tab:peak_memory} gives the measured training GPU memory.

\begin{table}[htbp]
\centering
\caption{Training memory for simulated cells with increasing peak counts. Settings: BF16, one cell per GPU, memory-efficient exact attention.}
\label{tab:peak_memory}
\small
\setlength{\tabcolsep}{7pt}
\begin{tabular}{lrrrrrr}
\toprule
Peaks per cell & 8,192 & 16,384 & 24,576 & 32,768 & 49,152 & 65,536 \\
\midrule
GPU memory (GB) & 29.2 & 40.4 & 55.0 & 69.5 & 98.7 & 127.8 \\
\bottomrule
\end{tabular}
\end{table}

The measured footprint is 69.5 GB for 32,768 peaks and 127.8 GB for 65,536 peaks, within nominal 80 GB and 141 GB GPU memory budgets, respectively. This demonstrates feasibility beyond the training cap in the profiled setting; it is not a runtime scaling experiment or a guarantee for arbitrary batch sizes and implementations. Memory-efficient exact attention avoids materializing the complete attention matrix but does not remove the quadratic number of pairwise peak interactions. Sparse, linear, or low-rank approximations may reduce arithmetic cost and remain directions for future work.
\FloatBarrier

\clearpage
\section*{NeurIPS Paper Checklist}

\begin{enumerate}

\item \begin{minipage}[t]{\linewidth}
{\bf Claims}
    \par\vspace{3pt} Question: Do the main claims made in the abstract and introduction accurately reflect the paper's contributions and scope?
    \par\vspace{3pt} Answer: \answerYes{}
    \par\vspace{3pt} Justification: The abstract and introduction state the main contributions of \method{}, including sequence-aware peak and gene encoding, multi-scale regulatory modeling, and evaluation on gene expression prediction and downstream regulatory tasks. These claims are supported by the method description and experimental results in Sections~\ref{sec:method} and~\ref{sec:exp}.
\end{minipage}

\item \begin{minipage}[t]{\linewidth}
{\bf Limitations}
    \par\vspace{3pt} Question: Does the paper discuss the limitations of the work performed by the authors?
    \par\vspace{3pt} Answer: \answerYes{}
    \par\vspace{3pt} Justification: Section~\ref{sec:conclusion} discusses encoder dependence, quadratic attention cost, paired-data requirements, restricted linking vocabularies, the scope of the spatial-context ablation, and BMMC-only multi-seed evaluation. Appendices~\ref{app:additional_ablations} and~\ref{app:efficiency} provide supporting details.
\end{minipage}

\item \begin{minipage}[t]{\linewidth}
{\bf Theory assumptions and proofs}
    \par\vspace{3pt} Question: For each theoretical result, does the paper provide the full set of assumptions and a complete (and correct) proof?
    \par\vspace{3pt} Answer: \answerNA{}
    \par\vspace{3pt} Justification: The paper does not present theoretical results or formal theorems. The mathematical formulations are used to describe the model architecture and training objectives.
\end{minipage}

\item \begin{minipage}[t]{\linewidth}
{\bf Experimental result reproducibility}
    \par\vspace{3pt} Question: Does the paper fully disclose all the information needed to reproduce the main experimental results of the paper to the extent that it affects the main claims and/or conclusions of the paper (regardless of whether the code and data are provided or not)?
    \par\vspace{3pt} Answer: \answerYes{}
    \par\vspace{3pt} Justification: The paper provides the model architecture, training objectives, data splits, evaluation protocols, hyperparameters, and compute details needed to reproduce the main experimental results. Additional implementation details are provided in the appendix.
\end{minipage}

\item \begin{minipage}[t]{\linewidth}
{\bf Open access to data and code}
    \par\vspace{3pt} Question: Does the paper provide open access to the data and code, with sufficient instructions to faithfully reproduce the main experimental results, as described in supplemental material?
    \par\vspace{3pt} Answer: \answerYes{}
    \par\vspace{3pt} Justification: The experiments are based on publicly available datasets and previously released baselines, all of which are cited in the paper. We provide detailed data processing, model configuration, and evaluation instructions in the appendix, and will release code and processed metadata to support reproducibility.
\end{minipage}

\item \begin{minipage}[t]{\linewidth}
{\bf Experimental setting/details}
    \par\vspace{3pt} Question: Does the paper specify all the training and test details (e.g., data splits, hyperparameters, how they were chosen, type of optimizer) necessary to understand the results?
    \par\vspace{3pt} Answer: \answerYes{}
    \par\vspace{3pt} Justification: The paper specifies the datasets, held-out sequence and cell-type splits, training objectives, optimizer settings, learning rates, batch sizes, epochs, and evaluation metrics. Full pre-training, fine-tuning, and evaluation details are reported in the appendix.
\end{minipage}

\item \begin{minipage}[t]{\linewidth}
{\bf Experiment statistical significance}
    \par\vspace{3pt} Question: Does the paper report error bars suitably and correctly defined or other appropriate information about the statistical significance of the experiments?
    \par\vspace{3pt} Answer: \answerYes{}
    \par\vspace{3pt} Justification: Clustering results report means and standard deviations over 10 seeds where available. Table~\ref{tab:gex_multiseed} reports five-seed BMMC PCC means and standard deviations. Enhancer--gene AUROC uses 95\% bootstrap confidence intervals; the motif benchmark uses gene-level confidence intervals. Other expression results remain single-run estimates, and unavailable PCA uncertainties are not imputed.
\end{minipage}

\item \begin{minipage}[t]{\linewidth}
{\bf Experiments compute resources}
    \par\vspace{3pt} Question: For each experiment, does the paper provide sufficient information on the computer resources (type of compute workers, memory, time of execution) needed to reproduce the experiments?
    \par\vspace{3pt} Answer: \answerYes{}
    \par\vspace{3pt} Justification: Section~\ref{sec:training_config} and Appendix~\ref{app:pretrain_details} give pretraining hardware and time. Appendix~\ref{app:efficiency} adds H200 inference timing and BF16 training-memory measurements versus peak count. These profiles do not cover end-to-end runtime for every downstream task.
\end{minipage}

\item \begin{minipage}[t]{\linewidth}
{\bf Code of ethics}
    \par\vspace{3pt} Question: Does the research conducted in the paper conform, in every respect, with the NeurIPS Code of Ethics \url{https://neurips.cc/public/EthicsGuidelines}?
    \par\vspace{3pt} Answer: \answerYes{}
    \par\vspace{3pt} Justification: The research uses publicly available or previously released biological datasets and does not involve deployment, decision-making systems, or collection of new human-subject data. The work conforms to the NeurIPS Code of Ethics.
\end{minipage}

\item \begin{minipage}[t]{\linewidth}
{\bf Broader impacts}
    \par\vspace{3pt} Question: Does the paper discuss both potential positive societal impacts and negative societal impacts of the work performed?
    \par\vspace{3pt} Answer: \answerYes{}
    \par\vspace{3pt} Justification: Section~\ref{sec:conclusion} discusses regulatory hypothesis generation and the risks of interpreting association scores as causal or clinical evidence. Experimental validation and oversight are required before therapeutic or sequence-engineering applications.
\end{minipage}

\item \begin{minipage}[t]{\linewidth}
{\bf Safeguards}
    \par\vspace{3pt} Question: Does the paper describe safeguards that have been put in place for responsible release of data or models that have a high risk for misuse (e.g., pre-trained language models, image generators, or scraped datasets)?
    \par\vspace{3pt} Answer: \answerNA{}
    \par\vspace{3pt} Justification: The work does not release a high-risk generative model, scraped dataset, or system with direct misuse potential. The proposed model is designed for biological representation learning and regulatory prediction.
\end{minipage}

\item \begin{minipage}[t]{\linewidth}
{\bf Licenses for existing assets}
    \par\vspace{3pt} Question: Are the creators or original owners of assets (e.g., code, data, models), used in the paper, properly credited and are the license and terms of use explicitly mentioned and properly respected?
    \par\vspace{3pt} Answer: \answerYes{}
    \par\vspace{3pt} Justification: Existing datasets, models, and baselines used in the paper are properly cited, including PBMC169K, BMMC, Openproblem, Bone, AlphaGenome, Borzoi, GET, EpiBERT, ChromFound, and other baselines. We use these assets according to their intended research use and do not redistribute restricted raw data.
\end{minipage}

\item \begin{minipage}[t]{\linewidth}
{\bf New assets}
    \par\vspace{3pt} Question: Are new assets introduced in the paper well documented and is the documentation provided alongside the assets?
    \par\vspace{3pt} Answer: \answerNA{}
    \par\vspace{3pt} Justification: The paper does not introduce a new dataset or benchmark asset. The proposed method and experimental protocols are described in the paper and appendix.
\end{minipage}

\item \begin{minipage}[t]{\linewidth}
{\bf Crowdsourcing and research with human subjects}
    \par\vspace{3pt} Question: For crowdsourcing experiments and research with human subjects, does the paper include the full text of instructions given to participants and screenshots, if applicable, as well as details about compensation (if any)?
    \par\vspace{3pt} Answer: \answerNA{}
    \par\vspace{3pt} Justification: The paper does not involve crowdsourcing experiments or direct interaction with human participants.
\end{minipage}

\item \begin{minipage}[t]{\linewidth}
{\bf Institutional review board (IRB) approvals or equivalent for research with human subjects}
    \par\vspace{3pt} Question: Does the paper describe potential risks incurred by study participants, whether such risks were disclosed to the subjects, and whether Institutional Review Board (IRB) approvals (or an equivalent approval/review based on the requirements of your country or institution) were obtained?
    \par\vspace{3pt} Answer: \answerNA{}
    \par\vspace{3pt} Justification: The paper uses publicly available, de-identified single-cell genomics datasets and does not conduct new human-subject data collection.
\end{minipage}

\item \begin{minipage}[t]{\linewidth}
{\bf Declaration of LLM usage}
    \par\vspace{3pt} Question: Does the paper describe the usage of LLMs if it is an important, original, or non-standard component of the core methods in this research? Note that if the LLM is used only for writing, editing, or formatting purposes and does \emph{not} impact the core methodology, scientific rigor, or originality of the research, declaration is not required.
    \par\vspace{3pt} Answer: \answerNA{}
    \par\vspace{3pt} Justification: LLMs are not used as an important, original, or non-standard component of the core method or experiments. Any use of writing assistance does not affect the scientific methodology, results, or claims.
\end{minipage}

\end{enumerate}

\end{document}